\documentclass{article}
\usepackage{typearea}
\typearea{12}
\usepackage[round]{natbib}
\usepackage{bm}
\usepackage{amsmath}
\usepackage{amssymb}
\usepackage{graphicx}
\def\vec#1{\mathbf{#1}}
\usepackage{mathtools}
\usepackage{comment}
\usepackage{wrapfig}
\usepackage{url}
\begin{document}

\title{Anomaly Detection with Inexact Labels}

\author{Tomoharu Iwata$^1$ \and
  Machiko Toyoda$^2$ \and
  Shotaro Tora$^2$ \and
  Naonori Ueda$^1$
}
\date{$^1$ NTT Communication Science Laboratories\\
  $^2$ NTT Software Innovation Center}

\maketitle

\begin{abstract}
  We propose a supervised anomaly detection method for data with inexact anomaly labels, where each label, which is assigned to a set of instances, indicates that at least one instance in the set is anomalous. Although many anomaly detection methods have been proposed, they cannot handle inexact anomaly labels. To measure the performance with inexact anomaly labels, we define the inexact AUC, which is our extension of the area under the ROC curve (AUC) for inexact labels. The proposed method trains an anomaly score function so that the smooth approximation of the inexact AUC increases while anomaly scores for non-anomalous instances become low. We model the anomaly score function by a neural network-based unsupervised anomaly detection method, e.g., autoencoders. The proposed method performs well even when only a small number of inexact labels are available by incorporating an unsupervised anomaly detection mechanism with inexact AUC maximization. Using various datasets, we experimentally demonstrate that our proposed method improves the anomaly detection performance with inexact anomaly labels, and outperforms existing unsupervised and supervised anomaly detection and multiple instance learning methods.
\end{abstract}

\section{Introduction}

Anomaly detection is an important machine learning task, 
which is a task to find the anomalous instances in a dataset.
Anomaly detection has been used
in a wide variety of applications~\citep{chandola2009anomaly,patcha2007overview,hodge2004survey},
such as
network intrusion detection for cyber-security~\citep{dokas2002data,yamanishi2004line},
fraud detection for credit cards~\citep{aleskerov1997cardwatch},
defect detection in industrial machines~\citep{fujimaki2005approach,ide2004eigenspace}
and disease outbreak detection~\citep{wong2003bayesian}.

Many unsupervised anomaly detection methods have been proposed~\citep{breunig2000lof,scholkopf2001estimating,liu2008isolation,sakurada2014anomaly}.
When anomaly labels, which indicate whether each instance is anomalous, are given,
the anomaly detection performance can be 
improved~\citep{singh2009ensemble,mukkamala2005model,rapaka2003intrusion,nadeem2016semi,gao2006novel,das2016incorporating,das2017incorporating}.
However, it is difficult to attach exact anomaly labels in some situations.
Consider such example in server system failure detection.
System operators often do not know the exact time of failures;
they only know that a failure occurred within a certain period of time.
In this case, anomaly labels can be attached to instances
in a certain period of time, in which non-anomalous instances might be included.
Another example is detecting anomalous IoT devices connected to a server.
Even when the server displays unusual behavior, we sometimes cannot identify
which IoT devices are anomalous.
In these situations, only inexact anomaly labels are available.

In this paper, we propose a supervised anomaly detection method for data with inexact anomaly 
labels.
An inexact anomaly label is attached to a set of instances,
indicating that at least one instance in the set is anomalous.
We call this set an inexact anomaly set.
First, we define an extension of the area under the ROC curve (AUC) for performance measurement with inexact labels,
which we call an {\it inexact AUC}.
Then we develop an anomaly detection method that maximizes the inexact AUC.
With the proposed method,
a function, which outputs an anomaly score given an instance, is modeled by the
reconstruction error with autoencoders, 
which are a successfully used neural network-based unsupervised anomaly detection 
method~\citep{sakurada2014anomaly,sabokrou2016video,chong2017abnormal,zhou2017anomaly}.
Note that the proposed method can use any unsupervised anomaly detection methods with learnable parameters 
instead of autoencoders, such as 
variational autoencoders~\citep{kingma2013auto},
energy-based models~\citep{zhai2016deep}, and
isolation forests~\citep{liu2008isolation}.
The parameters of the anomaly score function are trained so that
the anomaly scores for non-anomalous instances become low
while the smooth approximation of the inexact AUC becomes high.
Since our objective function is differentiable,
the anomaly score function can be estimated efficiently using
stochastic gradient-based optimization methods.

The proposed method performs well
even when only a few inexact labels are given
since it incorporates an unsupervised anomaly detection mechanism, 
which works without label information.
In addition, the proposed method is robust to class imbalance
since our proposed inexact AUC maximization is related to AUC maximization,
which achieved high performance on imbalanced data 
classification tasks~\citep{cortes2004auc}.
Class imbalance robustness is important for anomaly detection
since anomalous instances occur more rarely than non-anomalous instances.

\begin{figure}
\centering
\fbox{\includegraphics[width=15em]{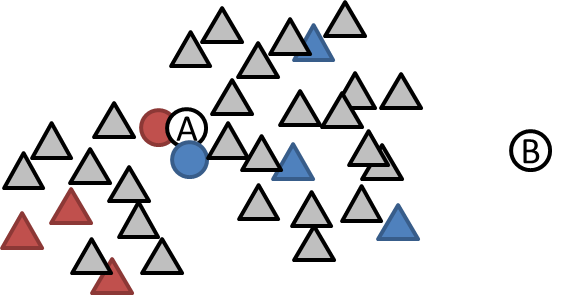}}
\caption{Example of anomalous (circle) and non-anomalous (triangle) instances, and inexact anomaly sets (red or blue) in an instance space. Instances with identical color (red or blue) are contained
in the same inexact anomaly set. White circles are test anomalous instances.}
\label{fig:example}
\end{figure}

Figure~\ref{fig:example} shows an example of 
anomalous and non-anomalous instances, and inexact anomaly sets in a two-dimensional instance space.
For unsupervised methods,
it is difficult to detect test anomalous instance `A'
since some instances are located around it.
Unsupervised methods consider that an instance is anomalous when
there are few instances around it.
Since supervised methods can use label information,
they can correctly detect test anomalous instance `A'.
However, they might not detect test anomalous instance `B' 
since there are no labeled instances near it.
In addition, with supervised methods that consider all the instances in inexact anomaly sets are anomaly,
non-anomalous instances around training instances in the inexact anomaly sets 
(colored triangles in Figure~\ref{fig:example})
would be misclassified as anomaly.
On the other hand,
the proposed method detects `A' using label information and
`B' by incorporating an unsupervised anomaly detection mechanism.
Also, it would not detect non-anomalous instances as anomaly 
since it can handle inexact information.

The remainder of the paper is organized as follows.
In Section~\ref{sec:related}, we briefly review related work.
In Section~\ref{sec:preliminaries}, we introduce AUC, 
which is the basis of the inexact AUC.
In Section~\ref{sec:proposed}, we present the inexact AUC,
define our task,
and propose our method for supervised anomaly detection using inexact labels.
In Section~\ref{sec:experiments}, we experimentally demonstrate the effectiveness
of our proposed method using various datasets by comparing with
existing anomaly detection and multiple instance learning methods.
Finally,
we present concluding remarks and
discuss future work in Section~\ref{sec:conclusion}.

\section{Related work}
\label{sec:related}

Inexact labels in classification tasks have been considered
in multiple instance learning methods~\citep{dietterich1997solving,maron1998framework,babenko2009visual,wu2015deep,cinbis2017weakly},
where labeled sets are given for training.
For a binary classification task,
a set is labeled negative if all the instances in it are negative,
and it is labeled positive if it contains at least one positive instance.
An advantage of the proposed method over existing multiple instance learning methods
is that the proposed method works well with a small number of inexact anomaly labels
using inexact AUC maximization and
incorporating an unsupervised anomaly detection mechanism,
where we exploit the characteristics of anomaly detection tasks.
Existing multiple instance learning methods are not robust to class imbalance~\citep{herrera2016multiple,carbonneau2018multiple}.

Anomaly detection is also called outlier detection~\citep{hodge2004survey}
or novelty detection~\citep{markou2003novelty}.
Many unsupervised methods have been proposed,
such as the local outlier factor~\citep{breunig2000lof},
one-class support vector machines~\citep{scholkopf2001estimating},
isolation forests~\citep{liu2008isolation},
and density estimation based 
methods~\citep{shewhart1931economic,eskin2000anomaly,laxhammar2009anomaly}.
However, these methods cannot use label information.
Although supervised anomaly detection methods
have been proposed to exploit
label information~\citep{nadeem2016semi,gao2006novel,das2016incorporating,das2017incorporating,Munawar_2017,pimentel2018generalized,akcay2018ganomaly,iwata2019supervised},
they cannot handle inexact anomaly labels.
A number of AUC maximization methods have been proposed~\citep{cortes2004auc,brefeld2005auc,ying2016stochastic,fujino2016semi,narasimhan2017support,sakai2018semi}
for training on class imbalanced data.
However, these methods do not consider inexact labels.

\section{Preliminaries: AUC}
\label{sec:preliminaries}

Let $\mathcal{X}$ be an instance space, and let
$p_{\mathrm{A}}$ and $p_{\mathrm{N}}$ be probability distributions over
anomalous and non-anomalous instances in $\mathcal{X}$.
Suppose that $a:\mathcal{X}\to \mathbb{R}$ is an anomaly score function,
and anomaly detection is carried out based on its sign:
\begin{align}
  \mathrm{sign}(a(\vec{x})-h),
\end{align}
where $\vec{x}$ is an instance and $h$ is a threshold.
The true positive rate (TPR) of anomaly score function $a(\vec{x})$
is the rate that it correctly classifies a random anomaly from $p_{\mathrm{A}}$
as anomalous,
\begin{align}
\mathrm{TPR}(h)=\mathbb{E}_{\vec{x}^{\mathrm{A}}
  \sim p_{\mathrm{A}}}[I(a(\vec{x}^{\mathrm{A}})>h)],
\end{align}
where $\mathbb{E}$ is the expectation and $I(\cdot)$ is the indicator function;
$I(A)=1$ if $A$ is true, and $I(A)=0$ otherwise.
The false positive rate (FPR) is the rate that it misclassifies a random 
non-anomalous instance from $p_{\mathrm{N}}$ as anomalous,
\begin{align}
  \mathrm{FPR}(h)=\mathbb{E}_{\vec{x}^{\mathrm{N}}\sim p_{\mathrm{N}}}[I(a(\vec{x}^{\mathrm{N}})>h)].
\end{align}
The ROC curve is the plot of $\mathrm{TPR}(h)$ as a function of 
$\mathrm{FPR}(h)$ with different threshold $h$. The area under this curve (AUC)~\citep{hanley1982meaning}
is computed as follows~\citep{dodd2003partial}:
\begin{align}
\mathrm{AUC}
=\int_{0}^{1}\mathrm{TPR}(\mathrm{FPR}^{-1}(s))ds
=\mathbb{E}_{\vec{x}^{\mathrm{A}}\sim p_{\mathrm{A}},\vec{x}^{\mathrm{N}}\sim p_{\mathrm{N}}}
[I(a(\vec{x}^{\mathrm{A}})>a(\vec{x}^{\mathrm{N}}))],
\label{eq:auc}
\end{align}
where $\mathrm{FPR}^{-1}(s)=\inf \{h\in\mathbb{R}|\mathrm{FPR}(h)\leq s\}$.
AUC is the rate where
a randomly sampled anomalous instance has a higher 
anomaly score than a randomly sampled non-anomalous instance.

Given sets of anomalous instances
$\mathcal{A}=\{\vec{x}^{\mathrm A}_{i}\}_{i=1}^{|\mathcal{A}|}$ 
drawn from $p_{\mathrm{A}}$ and 
non-anomalous instances 
$\mathcal{N}=\{\vec{x}^{\mathrm N}_{j}\}_{j=1}^{|\mathcal{N}|}$
drawn from $p_{\mathrm{N}}$,
an empirical AUC is calculated by
\begin{align}
\widehat{\mathrm{AUC}}=\frac{1}{|\mathcal{A}||\mathcal{N}|}
\sum_{\vec{x}_{i}^{\mathrm{A}}\in\mathcal{A}}\sum_{\vec{x}_{j}^{\mathrm{N}}\in\mathcal{N}}
I(a(\vec{x}_{i}^{\mathrm{A}})>a(\vec{x}_{j}^{\mathrm{N}})),
\label{eq:empauc}
\end{align}
where $|\mathcal{A}|$ represents the size of set $\mathcal{A}$.

\section{Proposed method}
\label{sec:proposed}

\subsection{Inexact AUC}

Let $\mathcal{B}=\{\vec{x}^{\mathrm{B}}_{i}\}_{i=1}^{|\mathcal{B}|}$ be a set of instances drawn from probability distribution $p_{\mathrm{S}}$,
where at least one instance is drawn from anomalous distribution $p_{\mathrm{A}}$,
and the other instances are drawn from non-anomalous distribution $p_{\mathrm{N}}$.
We define inexact true positive rate (inexact TPR)
as the rate where anomaly score function $a(\vec{x})$
classifies at least one instance in 
a random instance set from $p_{\mathrm{S}}$
as anomalous:
\begin{align}
\mathrm{iTPR}(h)
=\mathbb{E}_{\mathcal{B}\sim p_{\mathrm{S}}}
\left[I\left(
\bigvee_{\vec{x}_{i}^{\mathrm{B}}\in\mathcal{B}}
\left[
a(\vec{x}_{i}^{\mathrm{B}})>h)\right]
\right)
\right]
=\mathbb{E}_{\mathcal{B}\sim p_{\mathrm{S}}}
\left[I(\max_{x_{i}^{\mathrm{B}}\in\mathcal{B}}a(\vec{x}_{i}^{\mathrm{B}})>h)\right].
\label{eq:inexact_tpr}
\end{align}
We then define the inexact AUC
by the area under the curve of $\mathrm{iTPR}(h)$ as a function of 
$\mathrm{FPR}(h)$ with different threshold $h$ in
a similar way with the AUC (\ref{eq:auc}) as follows:
\begin{align}
\mathrm{iAUC}
=\int_{0}^{1}\mathrm{iTPR}(\mathrm{FPR}^{-1}(s))ds
=\mathbb{E}_{\mathcal{B}\sim p_{\mathrm{S}},\vec{x}^{\mathrm{N}}\sim p_{\mathrm N}}
[I(\max_{\vec{x}_{i}^{\mathrm{B}}\in\mathcal{B}}a(\vec{x}_{i}^{\mathrm{B}})>a(\vec{x}^{\mathrm{N}}))].
\label{eq:iauc}
\end{align}
Inexact AUC is the rate where at least one instance in 
a randomly sampled inexact anomaly set has a higher 
anomaly score than a randomly sampled non-anomalous instance.
When the label information is exact, 
i.e., every inexact anomaly set $\mathcal{B}$ contains only a single anomalous instance,
the inexact AUC (\ref{eq:iauc}) is equivalent to AUC (\ref{eq:auc}).
Therefore, the inexact AUC is a natural extension of the AUC for inexact labels.

Given a set of inexact anomaly sets 
$\mathcal{S}=\{\mathcal{B}_{k}\}_{k=1}^{|\mathcal{S}|}$,
where $\mathcal{B}_{k}=\{\vec{x}^{\mathrm B}_{ki}\}_{i=1}^{|\mathcal{B}_{k}|}$,
drawn from $p_{\mathrm{S}}$,
and a set of non-anomalous instances 
$\mathcal{N}=\{\vec{x}^{\mathrm N}_{j}\}_{j=1}^{|\mathcal{N}|}$
drawn from $p_{\mathrm{N}}$,
we calculate an empirical inexact AUC as follows:
\begin{align}
\widehat{\mathrm{iAUC}}=\frac{1}{|\mathcal{S}||\mathcal{N}|}
\sum_{\mathcal{B}_{k}\in\mathcal{S}}
\sum_{\vec{x}_{j}^{\mathrm{N}}\in\mathcal{N}}
I[\max_{\vec{x}^{\mathrm{B}}_{ki}\in\mathcal{B}_{k}}a(\vec{x}^{\mathrm{B}}_{ki})>a(\vec{x}_{j}^{\mathrm{N}})].
\label{eq:emp_inexact_auc}
\end{align}
The maximum operator has been widely used for
multiple instance learning methods~\citep{maron1998framework,andrews2003support,pinheiro2015image,zhu2017deep,feng2017deep,ilse2018attention}.
The proposed inexact AUC can evaluate score functions properly even with class imbalanced data
by incorporating the maximum operator into the AUC framework.

\subsection{Task}

Suppose that we are given a set of inexact anomaly sets
$\mathcal{S}$ and a set of non-anomalous instances $\mathcal{N}$ for training.
Our task is to estimate anomaly scores of test instances,
which are not included in the training data,
so that the anomaly score is high when the test instance is anomalous,
and low when it is non-anomalous.

\begin{table}[t]
\centering
\caption{Our notation}
\label{tab:notation}
\begin{tabular}{ll}
\hline
Symbol & Description \\ 
\hline
$\mathcal{S}$ & set of inexact anomaly sets, $\{\mathcal{B}_{k}\}_{k=1}^{|\mathcal{S}|}$\\
$\mathcal{B}_{k}$ & $k$th inexact anomaly set,
where at least one instance
is anomaly, $\{\vec{x}^{\mathrm B}_{ki}\}_{i=1}^{|\mathcal{B}_{k}|}$\\
$\mathcal{A}$ & set of anomalous instances, $\{\vec{x}^{\mathrm A}_{i}\}_{i=1}^{|\mathcal{A}|}$\\
$\mathcal{N}$ & set of non-anomalous instances, $\{\vec{x}^{\mathrm N}_{j}\}_{j=1}^{|\mathcal{N}|}$\\
$a(\vec{x})$ & anomaly score of instance $\vec{x}$\\
\hline
\end{tabular} 
\end{table}

\subsection{Anomaly scores}

For the anomaly score function, 
we use the following reconstruction error with deep autoencoders:
\begin{align}
  a(\vec{x};\bm{\theta}) = \parallel \vec{x} - g(f(\vec{x};\bm{\theta}_{\mathrm{f}});\bm{\theta}_{\mathrm{g}})\parallel^{2},
  \end{align}
where $f(\cdot;\bm{\theta}_{\mathrm{f}})$ is an encoder modeled by a neural network with parameters $\bm{\theta}_{\mathrm{f}}$,
$g(\cdot;\bm{\theta}_{\mathrm{g}})$ is a decoder modeled by a neural network with parameters
$\bm{\theta}_{\mathrm{g}}$,
and $\bm{\theta}=\{\bm{\theta}_{\mathrm{f}},\bm{\theta}_{\mathrm{g}}\}$
is the parameters of the anomaly score function.
The reconstruction error of an instance is likely to be low when 
instances similar to it often appear in the training data,
and the reconstruction error is likely to be high when
no similar instances are contained in the training data.
With the proposed method, we can use other anomaly score functions 
that are differentiable with respect to parameters,
such as Gaussian mixtures~\citep{eskin2000anomaly,an2015variational,suh2016echo,xu2018unsupervised},
variational autoencoders~\citep{kingma2013auto},
energy-based models~\citep{zhai2016deep}, and
isolation forests~\citep{liu2008isolation}.

\subsection{Objective function}

With the proposed method,
parameters $\bm{\theta}$ are trained
by minimizing the anomaly scores for non-anomalous instances
while maximizing the empirical inexact AUC (\ref{eq:emp_inexact_auc}).
To make the empirical inexact AUC differentiable with respect to the parameters,
we use sigmoid function
$\sigma(A-B)=\frac{1}{1+\exp(-(A-B))}$
instead of step function $I(A>B)$,
which is often used for a smooth approximation of the step function.
Then the objective function to be minimized is given:
\begin{align}
E = \frac{1}{|\mathcal{N}|}\sum_{\vec{x}_{j}^{\mathrm{N}}\in\mathcal{N}} a(\vec{x}_{j}^{\mathrm{N}})
-\lambda 
\frac{1}{|\mathcal{S}||\mathcal{N}|}
\sum_{\mathcal{B}_{k}\in\mathcal{S}}
\sum_{\vec{x}_{j}^{\mathrm{N}}\in\mathcal{N}}
\sigma\left(
\max_{\vec{x}^{\mathrm{B}}_{ki}\in\mathcal{B}_{k}}a(\vec{x}^{\mathrm{B}}_{ki})-a(\vec{x}_{j}^{\mathrm{N}})\right),
\label{eq:loss}
\end{align}
where $\lambda\geq 0$ is a hyperparameter that can be tuned using the inexact AUC 
on the validation data.
When there are no inexact anomaly sets or $\lambda=0$,
the second term becomes zero, 
and the first term on the non-anomalous instances remains with the objective function,
which is the same objective function with a standard autoencoder.
By the unsupervised anomaly detection mechanism of the first term in (\ref{eq:loss}),
the proposed method can detect anomalous instances
even when there are few inexact anomaly sets.
The computational complexity of calculating the objective function (\ref{eq:loss}) is
$O(|\mathcal{S}||\mathcal{B}|+|\mathcal{S}||\mathcal{N}|)$,
where $|\mathcal{B}|$ is the average number of instances in an inexact anomaly set,
the first term is for finding the maximum of anomaly scores in every inexact anomaly set,
and the second term is for calculating the difference of scores between inexact anomalous instances and
non-anomalous instances in the second term in (\ref{eq:loss}).

\section{Experiments}
\label{sec:experiments}

\subsection{Data}

We evaluated our proposed supervised anomaly detection method 
with a synthetic dataset and 
nine datasets used for unsupervised anomaly detection~\citep{campos2016evaluation}~\footnote{The datasets were obtained from \url{http://www.dbs.ifi.lmu.de/research/outlier-evaluation/DAMI/}}.

The synthetic dataset was generated from a two-dimensional Gaussian mixture model shown in 
Figure~\ref{fig:synth}(a,b).
The non-anomalous instances were generated from 
two unit-variance Gaussian distributions with mean at (-2,0) and (2,0),
as shown by blue triangles in Figure~\ref{fig:synth}(a).
The anomalous instances were generated from
a Gaussian distribution with mean (0,-1.5) with a small variance 
and a Gaussian distribution with mean (0,3) with a wide variance
as shown by red circles in Figure~\ref{fig:synth}(a).
The latter anomalous Gaussian was only used for test data, and 
it was not used for training and validation data
as shown in Figure~\ref{fig:synth}(b).
We generated 500 instances from the non-anomalous Gaussians, and 
200 instances from the anomalous Gaussians.

Table~\ref{tab:stats} shows the following values of the nine anomaly detection datasets:
the number of anomalous instances $|\mathcal{A}|$,
the number of non-anomalous instances $|\mathcal{N}|$,
anomaly ratio $\frac{|\mathcal{A}|}{|\mathcal{N}|}$,
and
the number of attributes $D$.
Each attribute was linearly normalized to range $[0,1]$,
and duplicate instances were removed.
The original datasets contained only exact anomaly labels~\citep{campos2016evaluation}.
We constructed inexact anomaly sets by randomly sampling non-anomalous instances 
and an anomalous instance for each set.

We used 70\% of the non-anomalous instances and ten inexact anomaly sets for training,
15\% of the non-anomalous instances and five inexact anomaly sets for validation,
and the remaining instances for testing.
The number of instances in an inexact anomaly set
was five with training and validation data, and one
with test data;
the test data contained only exact anomaly labels.
For each inexact anomaly set, we included an anomalous instance, and
the other instances were non-anomalous. 
For the evaluation measurement, we used AUC on test data. 
For each dataset,
we randomly generated ten sets of training, validation and test data,
and calculated the average AUC over the ten sets.

\begin{table}[t!]
\centering
\caption{Statistics of datasets used in our experiments. 
$|\mathcal{A}|$ is the number of anomalous instances,
$|\mathcal{N}|$ is the number of non-anomalous instances, and
$D$ is the number of attributes.}
\label{tab:stats}
\begin{tabular}{lrrrr}
\hline
Data & $|\mathcal{A}|$ & $|\mathcal{N}|$ & $\frac{|\mathcal{A}|}{|\mathcal{N}|}$ & $D$\\
\hline
Annthyroid & 350 & 6666 & 0.053 & 21 \\
Cardiotocography & 413 & 1655 & 0.250 & 21 \\
InternetAds & 177 & 1598 & 0.111 & 1555 \\
KDDCup99 & 246 & 60593 & 0.004 & 79 \\
PageBlocks & 258 & 4913 & 0.053 & 10 \\
Pima & 125 & 500 & 0.250 & 8 \\
SpamBase & 697 & 2788 & 0.250 & 57 \\
Waveform & 100 & 3343 & 0.030 & 21 \\
Wilt & 93 & 4578 & 0.020 & 5 \\
\hline
\end{tabular}
\end{table}

\subsection{Comparing methods}

We compared our proposed method with the following 11 methods:
LOF, OSVM, IF, AE, KNN, SVM, RF, NN, MIL, SIF and SAE.
LOF, OSVM, IF and AE are unsupervised anomaly detection methods, 
where attribute $\vec{x}$ is used for calculating the anomaly score,
but the label information is not used for training.
KNN, SVM, RF, NN, MIL, SIF, SAE and our proposed method 
are supervised anomaly detection methods,
where both the attribute $\vec{x}$ and the label information are used.
Since KNN, SVM, RF, NN, SIF and SAE cannot handle inexact labels,
they assume that all the instances in the inexact anomaly sets are anomalous.
For hyperparameter tuning, we used the AUC scores on the validation data with
LOF, OSVM, IF, AE, KNN, SVM, RF, NN, SIF and SAE,
and inexact AUC with MIL and our proposed method.
We used the scikit-learn implementation~\citep{pedregosa2011scikit} with 
LOF, OSVM, IF, KNN, SVM, RF and NN.

{\bf LOF}, which is the local outlier factor method~\citep{breunig2000lof},
unsupervisedly detects anomalies
based on the degree of isolation from the surrounding neighborhood.
The number of neighbors was tuned from $\{1,3,5,15,35\}$ using the validation data.

{\bf OSVM} is the one-class support vector machine~\citep{scholkopf2001estimating},
which is an extension of the support vector machine (SVM) for unlabeled data.
OSVM finds the maximal margin hyperplane, which 
separates the given non-anomalous data from the origin by embedding them in a high-dimensional space 
by a kernel function.
We used the RBF kernel,
its kernel hyperparameter was tuned from $\{10^{-3},10^{-2},10^{-1},1\}$,
and hyperparameter $\nu$ was tuned from $\{10^{-3},5\times10^{-3},10^{-2},5\times10^{-2},10^{-1},0.5,1\}$.

{\bf IF} is the isolation forest method~\citep{liu2008isolation}, which is a tree-based
unsupervised anomaly detection scheme. IF isolates anomalies by randomly selecting an attribute
and randomly selecting a split value between the maximum and minimum values of the selected attribute.
The number of base estimators was chosen from $\{1,5,10,20,30\}$.

{\bf AE} 
calculates the anomaly score by the reconstruction error with the autoencoder,
which is also used with the proposed method.
We used the same parameter setting with the proposed method for AE,
which is described in the next subsection.
Although the model of the proposed method with $\lambda=0$ is the same with that of AE,
early stopping criteria were different, where the proposed method used the inexact AUC,
and AE used the AUC.

{\bf KNN} is the $k$-nearest neighbor method,
which classifies instances based on the votes of neighbors.
The number of neighbors was selected from $\{1,3,5,15\}$.

{\bf SVM} is a support vector machine~\citep{scholkopf2002learning},
which is a kernel-based binary classification method.
We used the RBF kernel,
and the kernel hyperparameter was tuned from $\{10^{-3},10^{-2},10^{-1},1\}$.

{\bf RF} is the random forest method~\citep{breiman2001random},
which is a meta estimator that fits a number of decision tree classifiers.
The number of trees was chosen from $\{5,10,20,30\}$.

{\bf NN} is a feed-forward neural network classifier.
We used three layers with rectified linear unit (ReLU) activation,
where the number of hidden units was selected from $\{5,10,50,100\}$.

{\bf MIL} is a multiple instance learning method based on an autoencoder,
which is trained by maximizing the inexact AUC.
We used the same parameter setting with the proposed method for the autoencoder.
The proposed method with $\lambda=\infty$ corresponds to MIL.

{\bf SIF} is a supervised anomaly detection method based on the isolation
forest~\citep{das2017incorporating},
where the weights of the isolation forest are adjusted by maximizing the AUC.

{\bf SAE} is a supervised anomaly detection method based on an autoencoder,
where the neural networks are learned by minimizing the reconstruction error
while maximizing the AUC.
We used the same parameter setting with the proposed method for the autoencoder.

\subsection{Settings of the proposed method}

We used three-layer feed-forward neural networks for the encoder and decoder, where
the hidden unit size was 128,
and the output layer of the encoder and the input layer of the decoder was 16.
Hyperparameter $\lambda$ was selected from $\{0,10^{-3},10^{-2},10^{-1},1,10,10^{2},10^{3}\}$
using the inexact AUC on the validation data.
The validation data were also used for early stopping, where the maximum number of training epochs was 1000.
We optimized the neural network parameters using ADAM~\citep{kingma2014adam}
with learning rate $10^{-3}$,
where we randomly sampled eight inexact anomaly sets and 128 non-anomalous instances for each batch.
We implemented all the methods based on PyTorch~\citep{paszke2017automatic}.

\subsection{Results}

\begin{figure*}[t!]
\centering
{\tabcolsep=-0.6em
\begin{tabular}{ccc}
\includegraphics[height=11em]{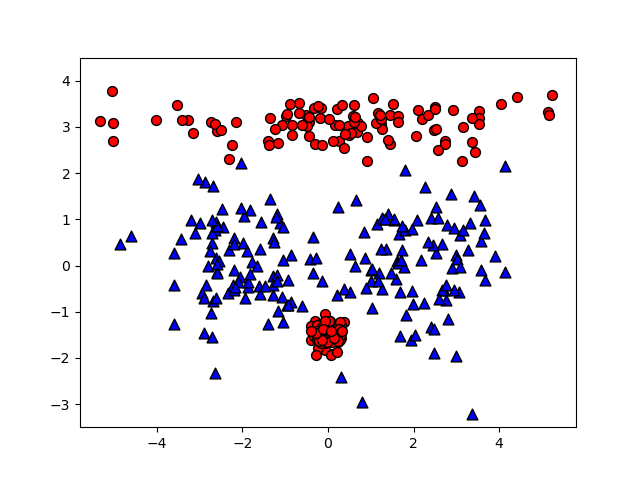}
&
\includegraphics[height=11em]{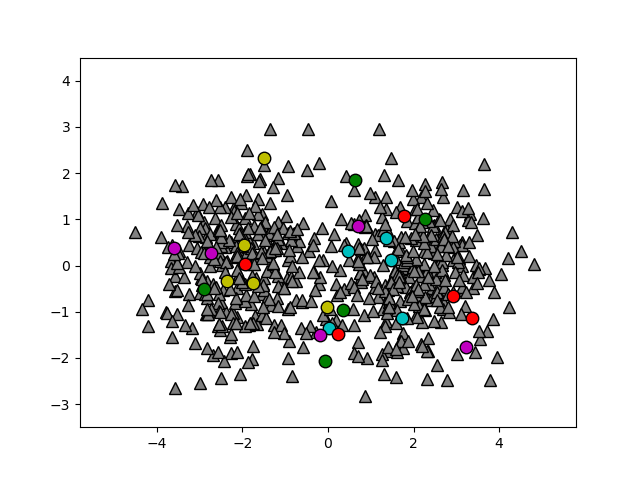}
&
\includegraphics[height=11em]{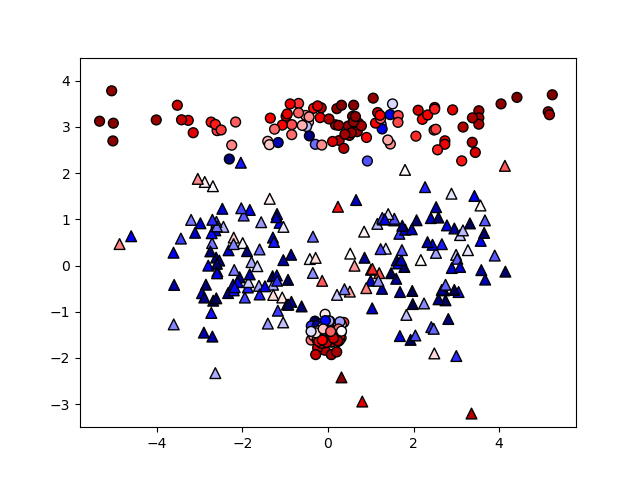}
\\
(a) Test data with true labels 
 & 
(b) Training data
& 
(c) AE \\
\includegraphics[height=11em]{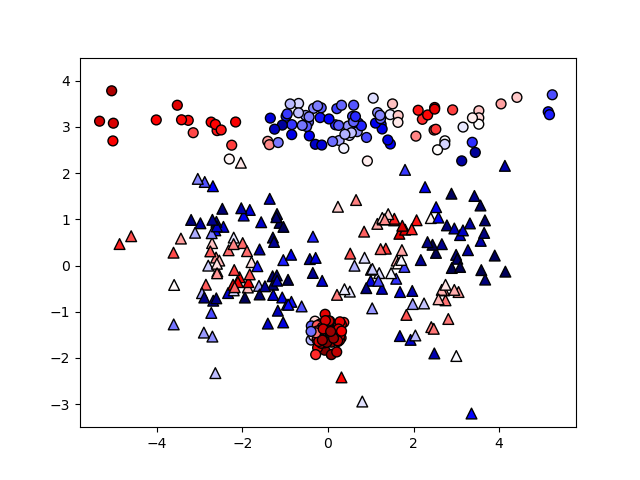}
&
\includegraphics[height=11em]{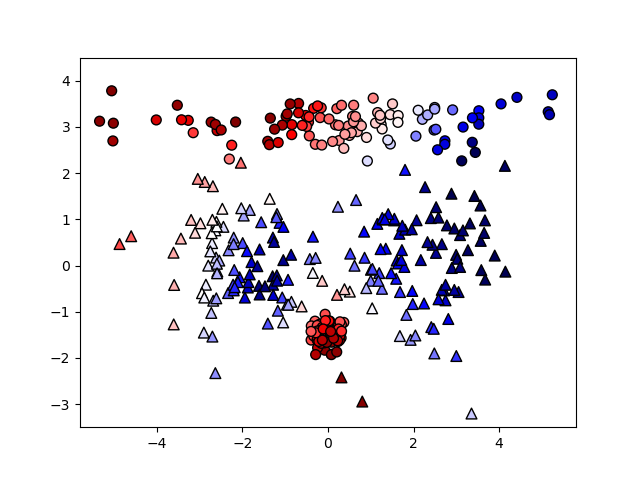}
&
\includegraphics[height=11em]{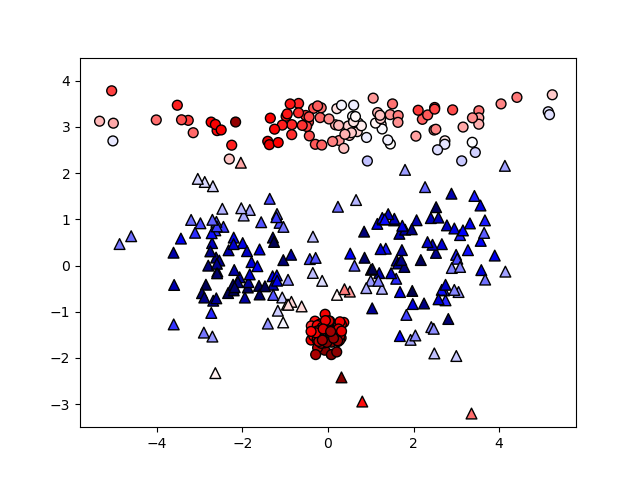}
\\
(d) SAE & (e) MIL & (f) Ours \\

\multicolumn{3}{c}{\includegraphics[width=18em]{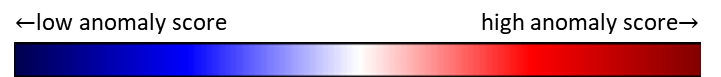}}
\\
\end{tabular}}
\caption{Synthetic dataset and the estimated anomaly scores in the two-dimensional instance space. (a) Test data: anomalous instances are represented by red circles and non-anomalous instances are represented by blue triangles. (b) Training data: instances in the same inexact anomalous set are represented by circles with identical color, and non-anomalous instances are represented by gray triangles. Note that anomalous instances at the top in the test data are not contained in the training data. (c--f) Estimated anomaly scores by the AE (c), SAE (d), MIL (e) and the proposed method (f). The shape indicates the true label (circle: anomalous, triangle: non-anomalous), the color indicates the estimated anomaly score; the darker red indicates the higher anomaly score, and the darker blue indicates lower anomaly score.}
\label{fig:synth}
\end{figure*}

\begin{figure*}[t!]
\centering
\begin{tabular}{cc}
\includegraphics[width=15em]{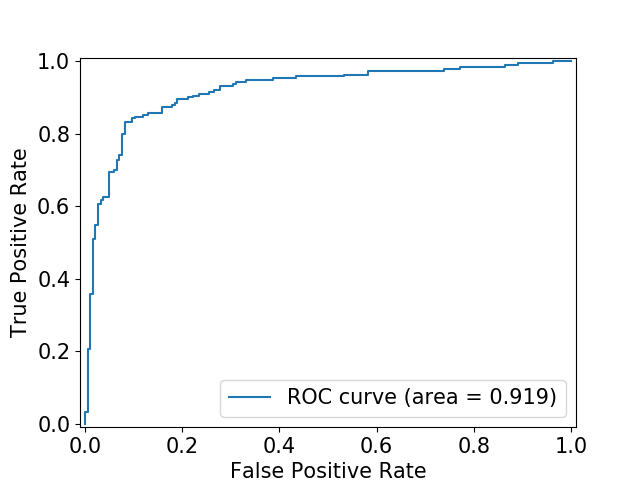}
&
\includegraphics[width=15em]{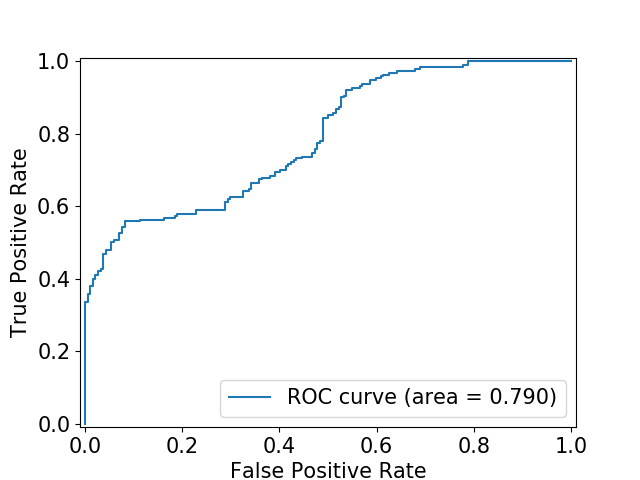}
\\
(a) AE & (b) SAE \\
\includegraphics[width=15em]{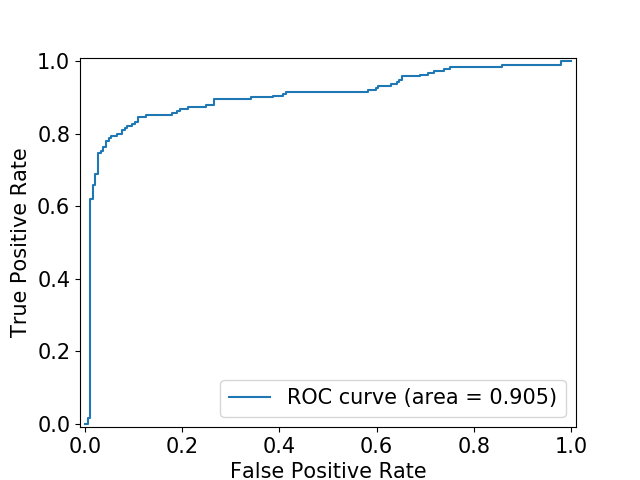}
&
\includegraphics[width=15em]{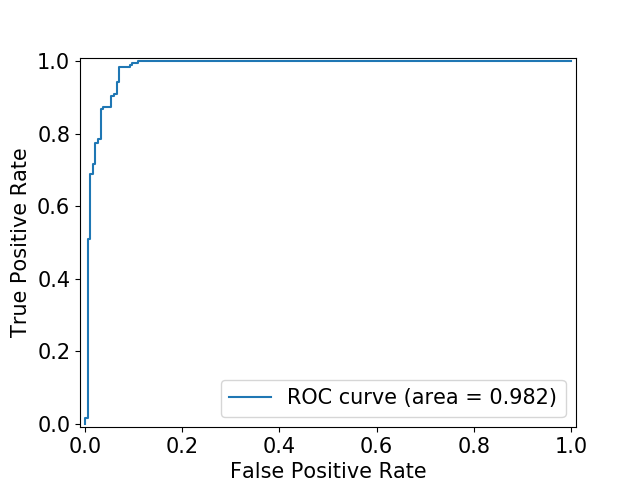}
\\
(c) MIL & (d) Ours\\
\end{tabular}
\caption{ROC curve and AUC on the test synthetic dataset by (a) AE, (b) SAE, (c) MIL, and (d) the proposed method. X-axis is the false positive rate, and y-axis is the true positive rate.}
\label{fig:roc}
\end{figure*}

Figure~\ref{fig:synth} shows the estimated anomaly scores by AE (c), SAE (d), MIL (e) and the proposed method (f) on the synthetic dataset.
Figure~\ref{fig:roc} shows the ROC curve and test AUC
by the AE (a), SAE (b), MIL (c) and the proposed method (d) on the synthetic dataset.
The test AUCs were 0.919 with AE, 0.790 with SAE, 0.905 with MIL, 
and 0.982 with the proposed method.
The AE successfully 
gave relatively high anomaly scores to the test anomalous instances at the top in 
Figure~\ref{fig:synth}(c).
However, since the AE is an unsupervised method and cannot use label information,
some anomalous instances at the bottom center were misclassified as non-anomaly.
The SAE is a supervised method, therefore the anomalous instances at the bottom center
in Figure~\ref{fig:synth}(d)
were identified as anomaly more appropriately than the AE.
However, since the SAE cannot handle inexact labels,
some test non-anomalous instances, that were located around 
non-anomalous instances in the training inexact anomaly sets,
were falsely classified as anomaly.
The MIL correctly gave lower anomaly scores to test non-anomalous instances
by handling inexact labels than the SAE 
in Figure~\ref{fig:synth}(e).
However, the MIL failed to correctly give high anomaly scores to
unseen anomalous instances at the top.
On the other hand, because the proposed method is trained by
minimizing the anomaly scores for non-anomalous instances 
while maximizing the inexact AUC,
the proposed method succeeded to detect 
unseen anomalous instances at the top as well as
anomalous instances at the bottom center, 
and correctly classified test non-anomalous instances as non-anomaly
in Figure~\ref{fig:synth}(f).

Table~\ref{tab:auc10_5}
shows AUC on the nine anomaly detection datasets with ten inexact anomaly sets and five instances per set.
Our proposed method achieved the highest AUC in most cases.
Since the number of supervised labels was small, 
the performance of the supervised methods, KNN, SVM, RF and MIL, was not high.
The proposed method outperformed them by incorporating an unsupervised method (AE) in a supervised framework.
SIF and SAE also used both unsupervised and supervised anomaly detection frameworks.
However, the performance was worse than the proposed method
because SIF and SAE cannot handle inexact labels.
Although MIL can handle inexact labels,
AUC with MIL was low since it does not have an unsupervised training mechanism, 
i.e., it does not minimize anomaly scores for non-anomalous instances.
The average computational time for training the proposed method was
1.0, 0.4, 1.4, 8.1, 0.7, 0.2, 0.5, 0.6 and 0.7 minutes with
Annthyroid, Cardiotocography, InternetAds, KDDCup99, PageBlocks,
Pima, SpamBase, Waveform and Wilt datasets, respectively,
on computers with 2.60GHz CPUs.

\begin{table*}[t!]
\centering
\caption{AUC on nine anomaly detection datasets with ten inexact anomaly sets and five instances per set. Values in bold typeface are not statistically different (at 5\% level) from the best performing method in each dataset according to a paired t-test. The Average column shows the average AUC over all datasets, and the value in bold indicates the best average AUC.}
\label{tab:auc10_5}
  \begin{tabular}{l|rrrrr}
    \hline
 &Annthyroid &Cardiotocography &InternetAds &KDDCup99 &PageBlocks \\
    \hline LOF & 0.652 & 0.544 & 0.728 & 0.576 & 0.754 \\
 OSVM & 0.525 & {\bf 0.845} & 0.814 & 0.974 & 0.877 \\
 IF & 0.768 & {\bf 0.809} & 0.549 & 0.973 & {\bf 0.924} \\
 AE & 0.754 & 0.768 & {\bf 0.839} & {\bf 0.995} & {\bf 0.915} \\
 KNN & 0.546 & 0.639 & 0.602 & 0.804 & 0.672 \\
 SVM & {\bf 0.751} & {\bf 0.725} & {\bf 0.864} & 0.708 & 0.599 \\
 RF & {\bf 0.868} & {\bf 0.806} & 0.622 & 0.895 & 0.862 \\
 NN & 0.622 & 0.702 & 0.783 & 0.975 & 0.462 \\
 MIL & 0.590 & {\bf 0.801} & 0.824 & 0.714 & 0.609 \\
 SIF & {\bf 0.829} & {\bf 0.843} & 0.622 & 0.992 & {\bf 0.932} \\
 SAE & {\bf 0.836} & 0.768 & 0.832 & 0.924 & {\bf 0.926} \\
 Ours  & {\bf 0.867} & {\bf 0.846} & 0.828 & {\bf 0.992} & {\bf 0.914}\\
    \hline
  \end{tabular}
  
  \begin{tabular}{l|rrrr|r}
    \hline
&Pima &SpamBase &Waveform &Wilt &Average \\
    \hline
 LOF & 0.601 & 0.546 & {\bf 0.680} & 0.709 & 0.643 \\
 OSVM & 0.686 & 0.639 & 0.622 & 0.571 & 0.728 \\
 IF & {\bf 0.714} & 0.703 & 0.660 & 0.617 & 0.746 \\
 AE & 0.678 & 0.757 & {\bf 0.671} & {\bf 0.895} & 0.808 \\
 KNN & 0.536 & 0.617 & 0.627 & 0.557 & 0.622 \\
 SVM & 0.495 & 0.573 & {\bf 0.729} & 0.665 & 0.679 \\
 RF & 0.649 & {\bf 0.751} & {\bf 0.711} & 0.774 & 0.771 \\
 NN & 0.396 & {\bf 0.782} & {\bf 0.724} & 0.619 & 0.674 \\
 MIL & 0.670 & 0.660 & {\bf 0.640} & 0.474 & 0.665 \\
 SIF & {\bf 0.706} & {\bf 0.808} & {\bf 0.723} & 0.703 & 0.795 \\
 SAE & 0.662 & {\bf 0.765} & {\bf 0.728} & {\bf 0.863} & 0.812 \\
 Ours  & {\bf 0.713} & {\bf 0.791} & {\bf 0.746} & {\bf 0.895} & {\bf 0.844}\\
    \hline
  \end{tabular}
\end{table*}  

\begin{figure}[t!]
  \centering
  {\tabcolsep=1em
  \begin{tabular}{cc}
    \includegraphics[height=12em]{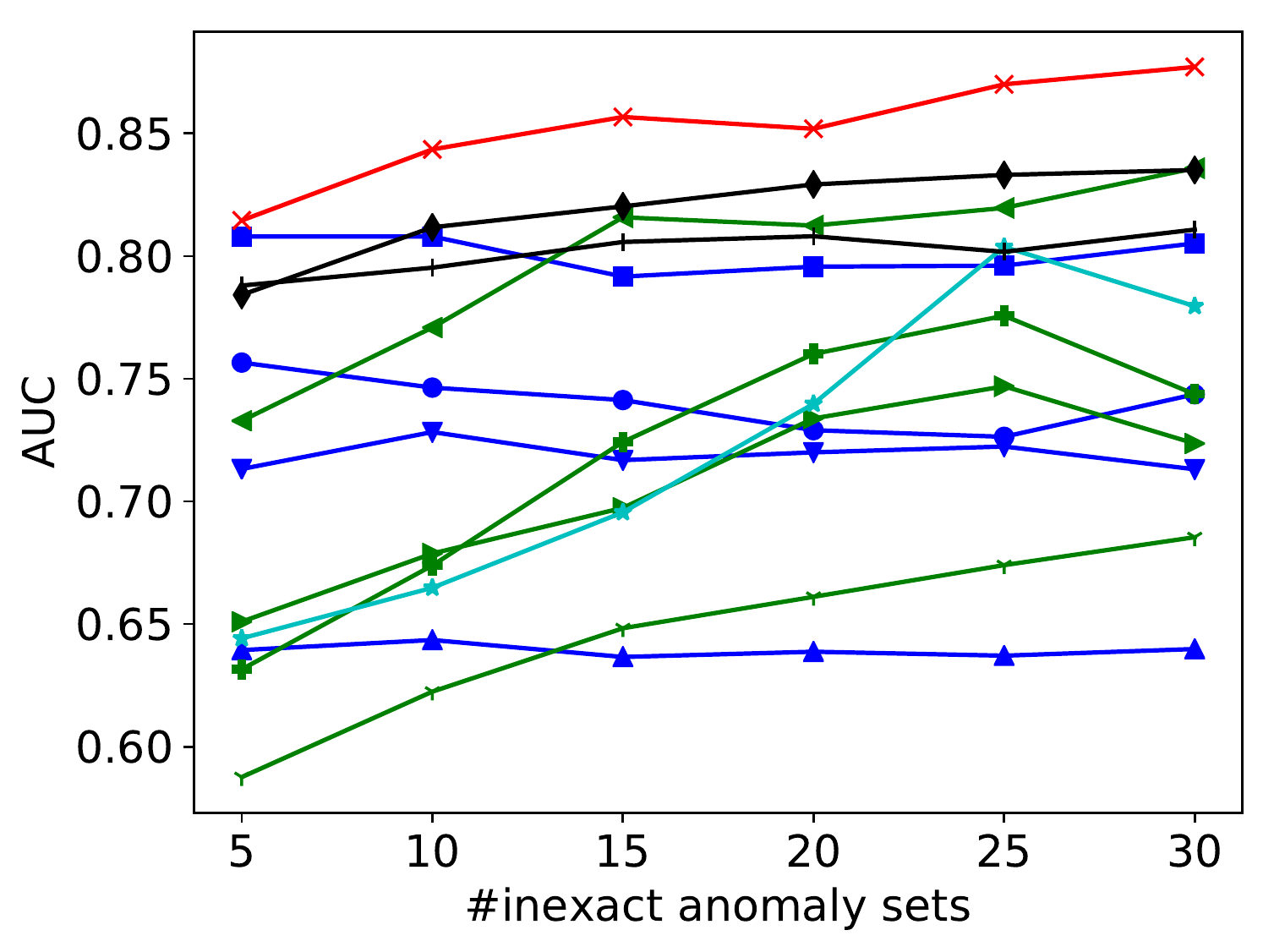}&
    \includegraphics[height=12em]{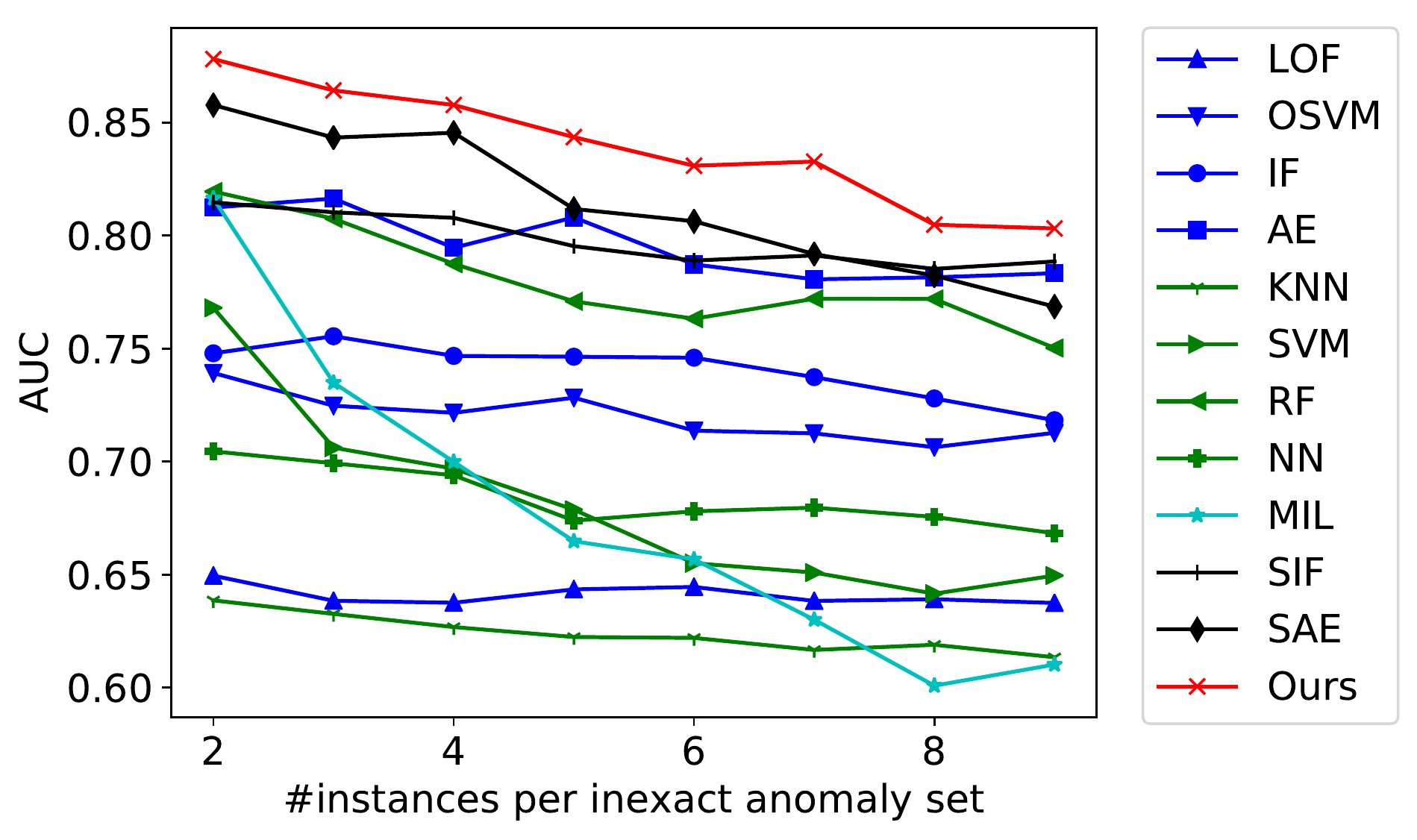}\\
    (a) number of inexact anomaly sets &
    (b) number of instances per inexact anomaly set\\
  \end{tabular}}
    \caption{AUC averaged over the nine anomaly detection datasets (a) with different numbers of training inexact anomaly sets and five instances per inexact anomaly set, and (b) with different numbers of instances per inexact anomaly set and ten training inexact anomaly sets.}
    \label{fig:auc}  
\end{figure}

Figure~\ref{fig:auc} shows test AUCs averaged over the nine anomaly detection datasets
by changing the number of training inexact anomaly sets (a), and
by changing the number of instances per inexact anomaly set (b).
The proposed method achieved the best performance in all cases.
As the number of training inexact anomaly sets increased,
the performance with supervised methods was improved.
As the number of instances per inexact anomaly set increased,
AUC was decreased
since the rate of non-anomalous instances in an inexact anomaly set increased.
AUC with unsupervised methods also decreased
since they used inexact anomaly sets in the validation data.

   \begin{figure*}[t!]
    \centering
   \begin{tabular}{ccc}
\includegraphics[width=11em]{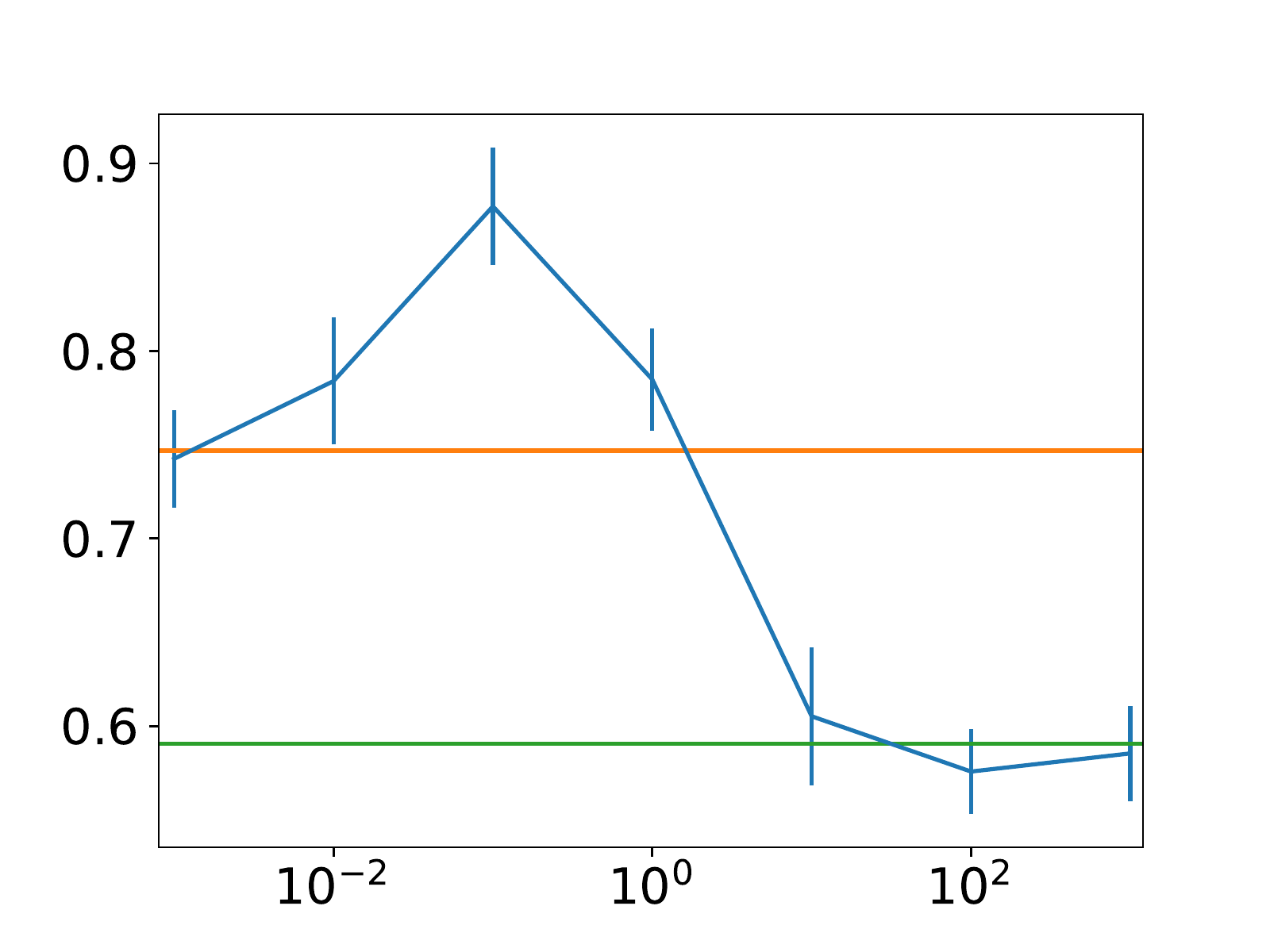} &
\includegraphics[width=11em]{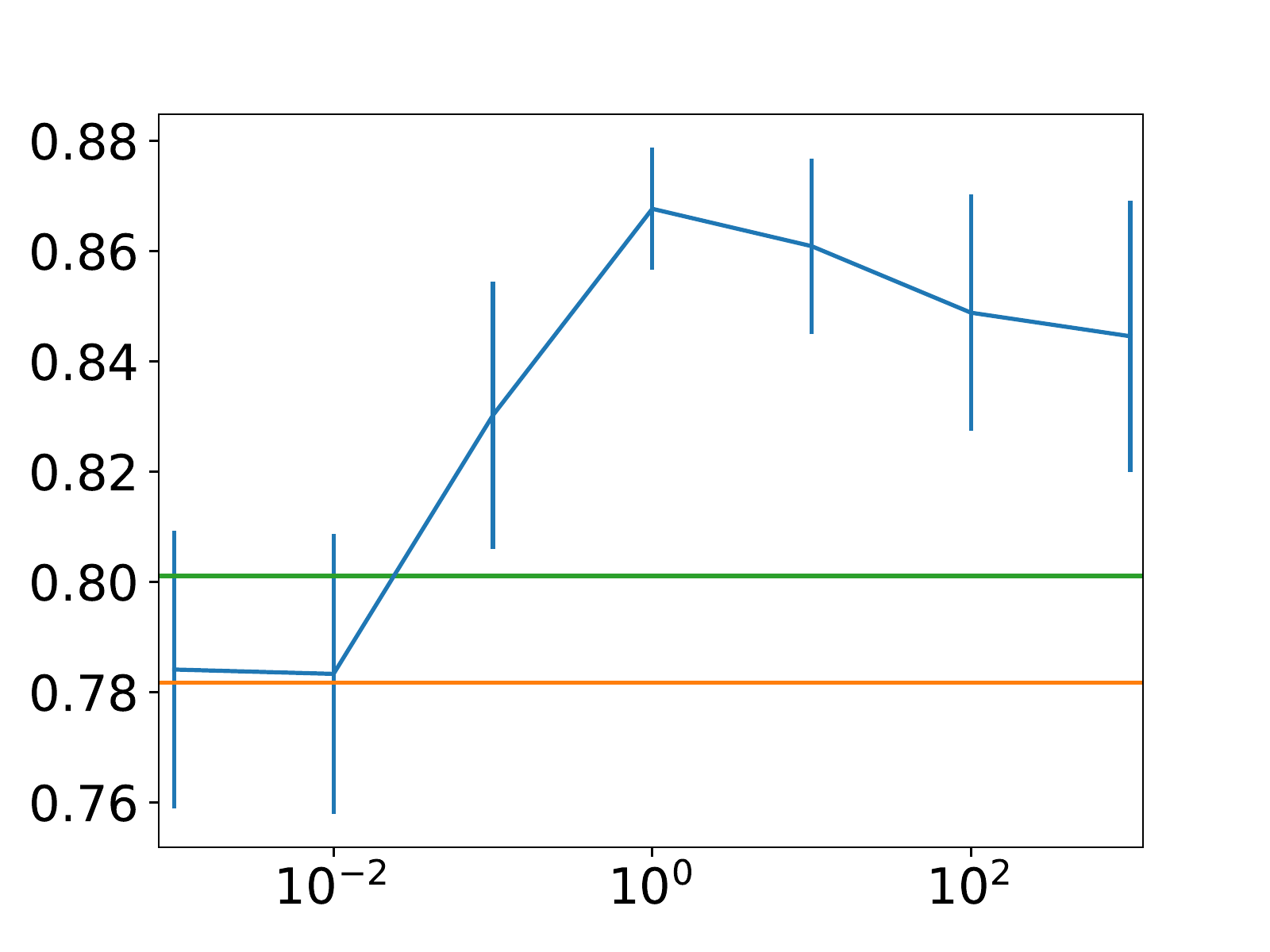} &
\includegraphics[width=11em]{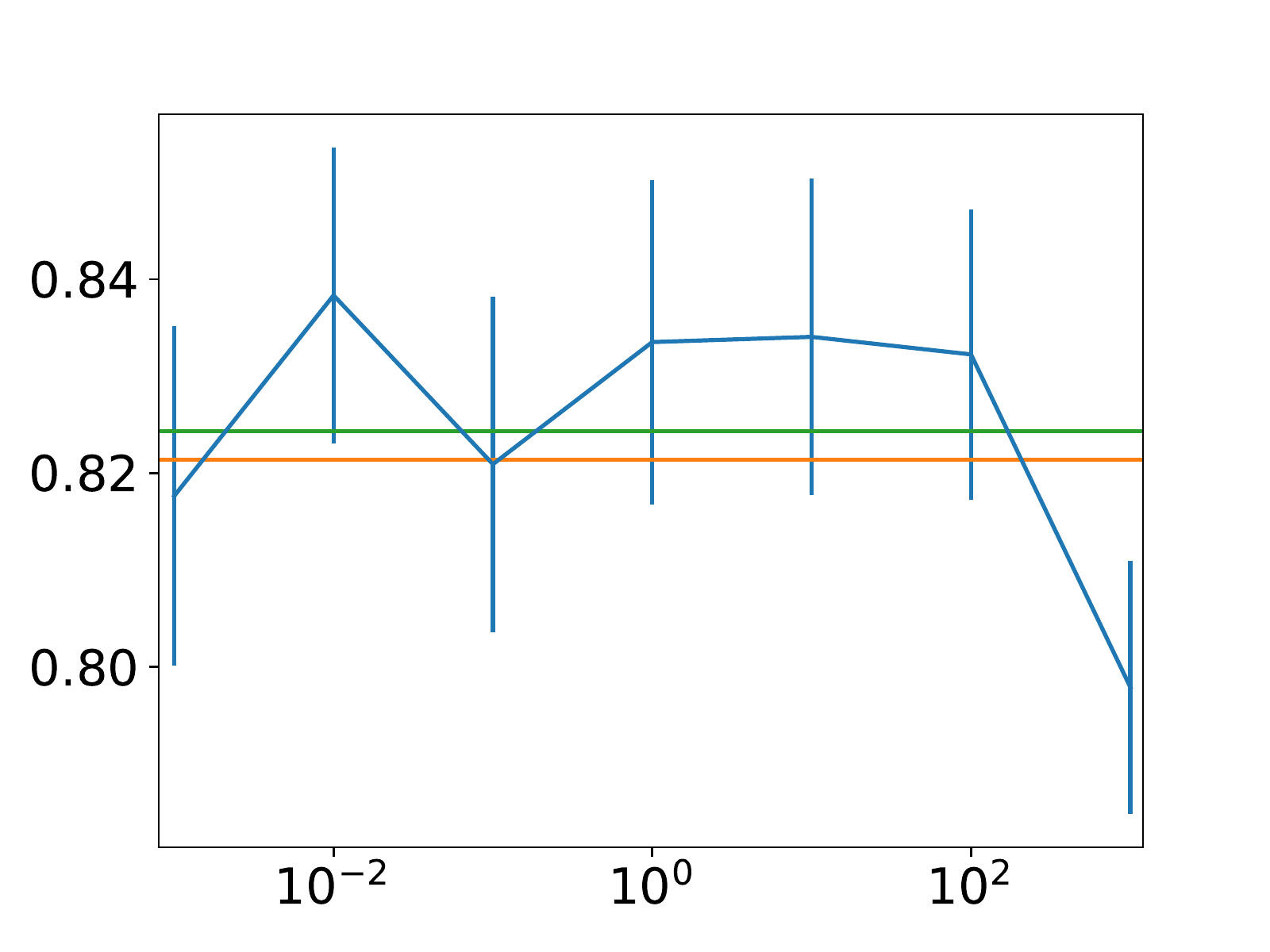} \\
Annthyroid&
Cardiotocography&
InternetAds\\
\includegraphics[width=11em]{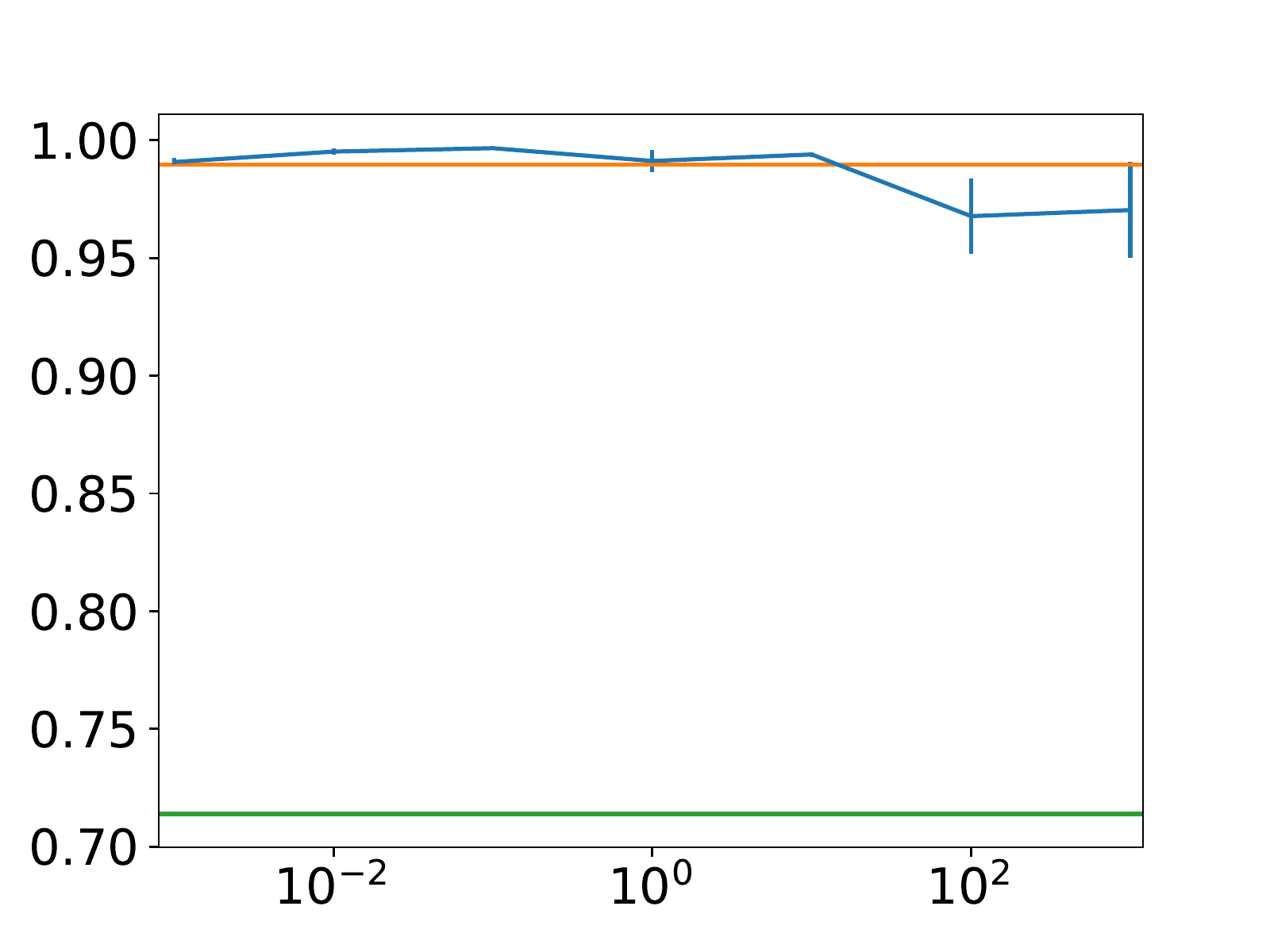} &
\includegraphics[width=11em]{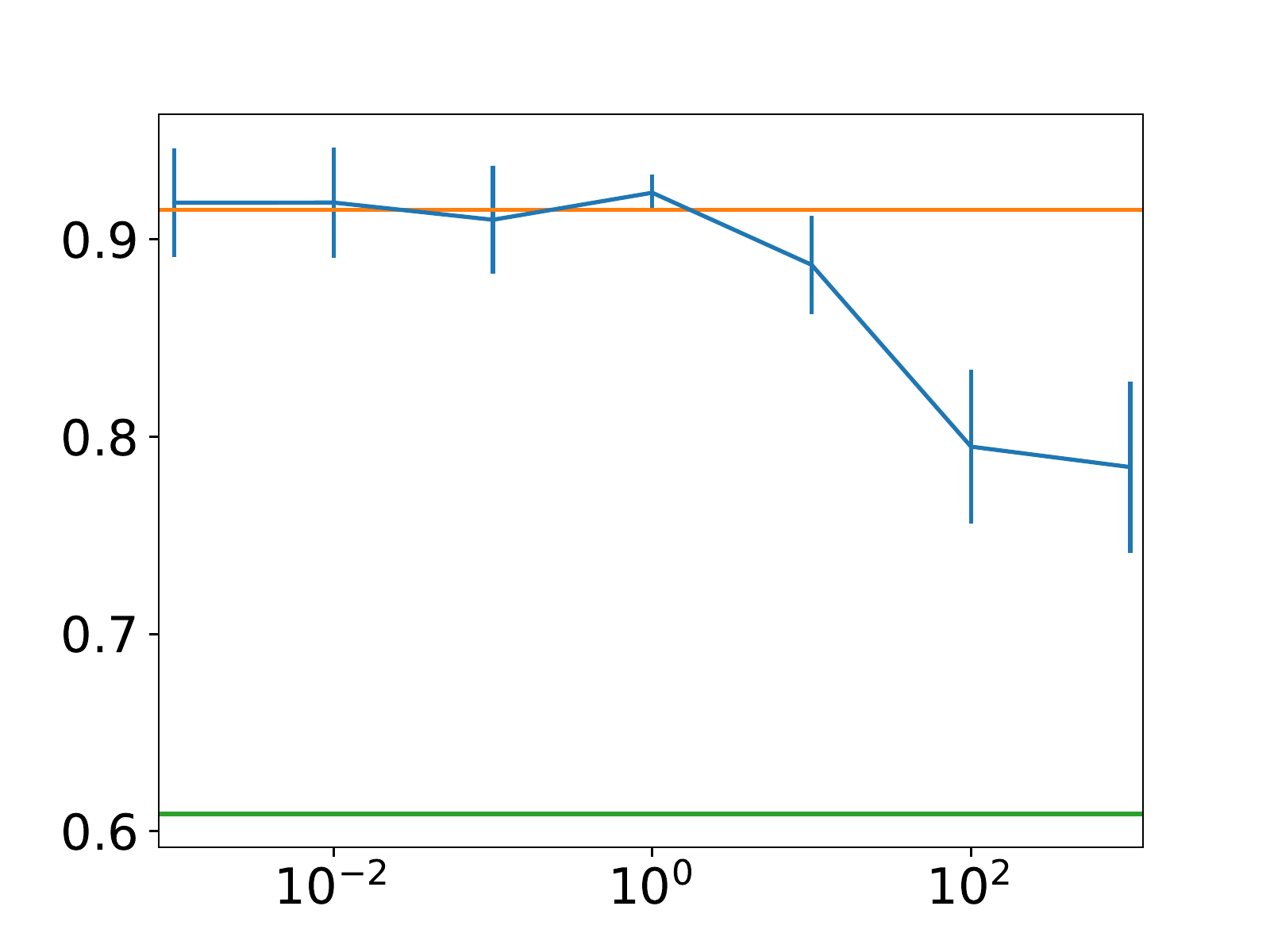} &
\includegraphics[width=11em]{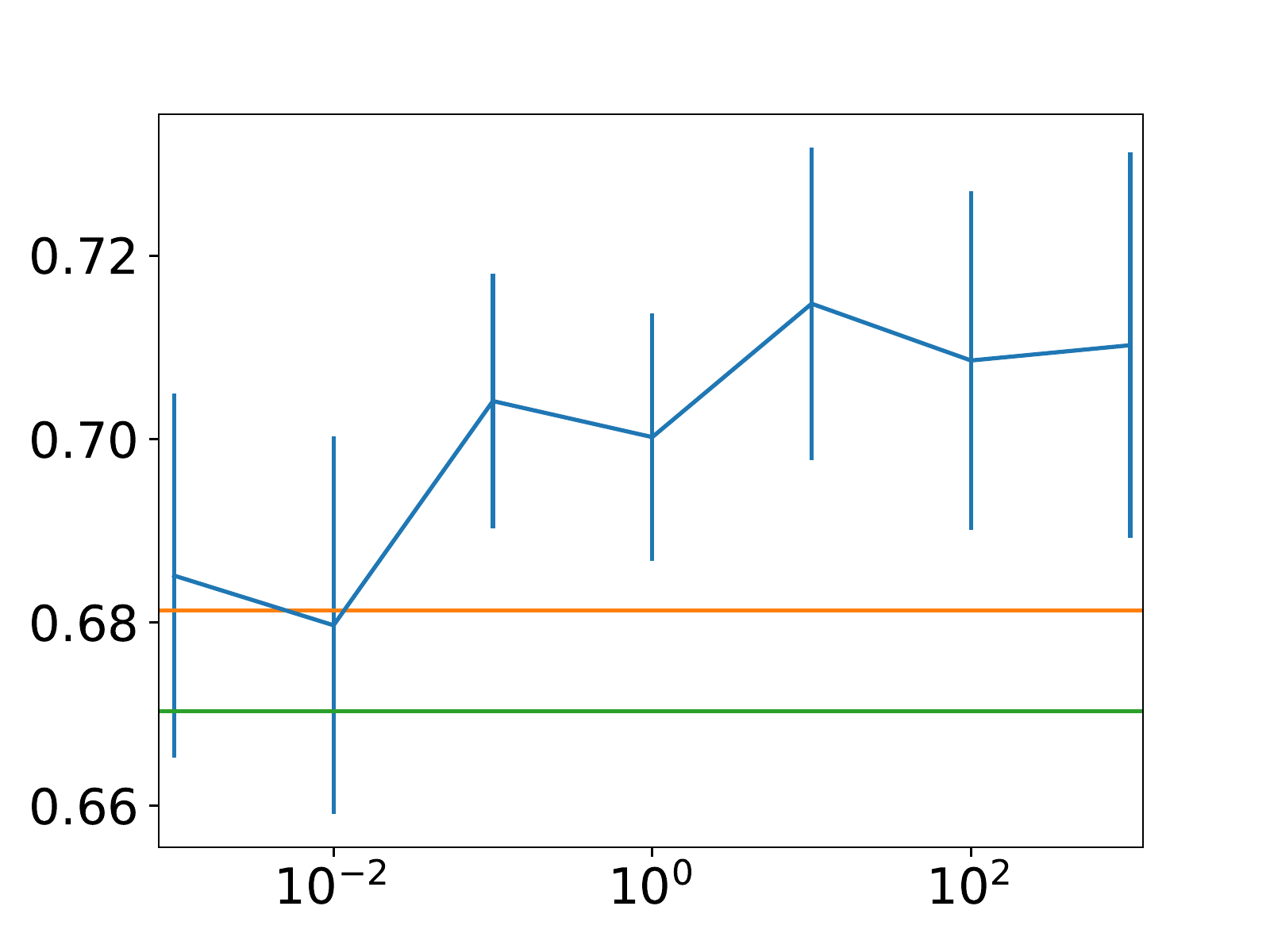} \\
KDDCup99&
PageBlocks &
Pima \\
\includegraphics[width=11em]{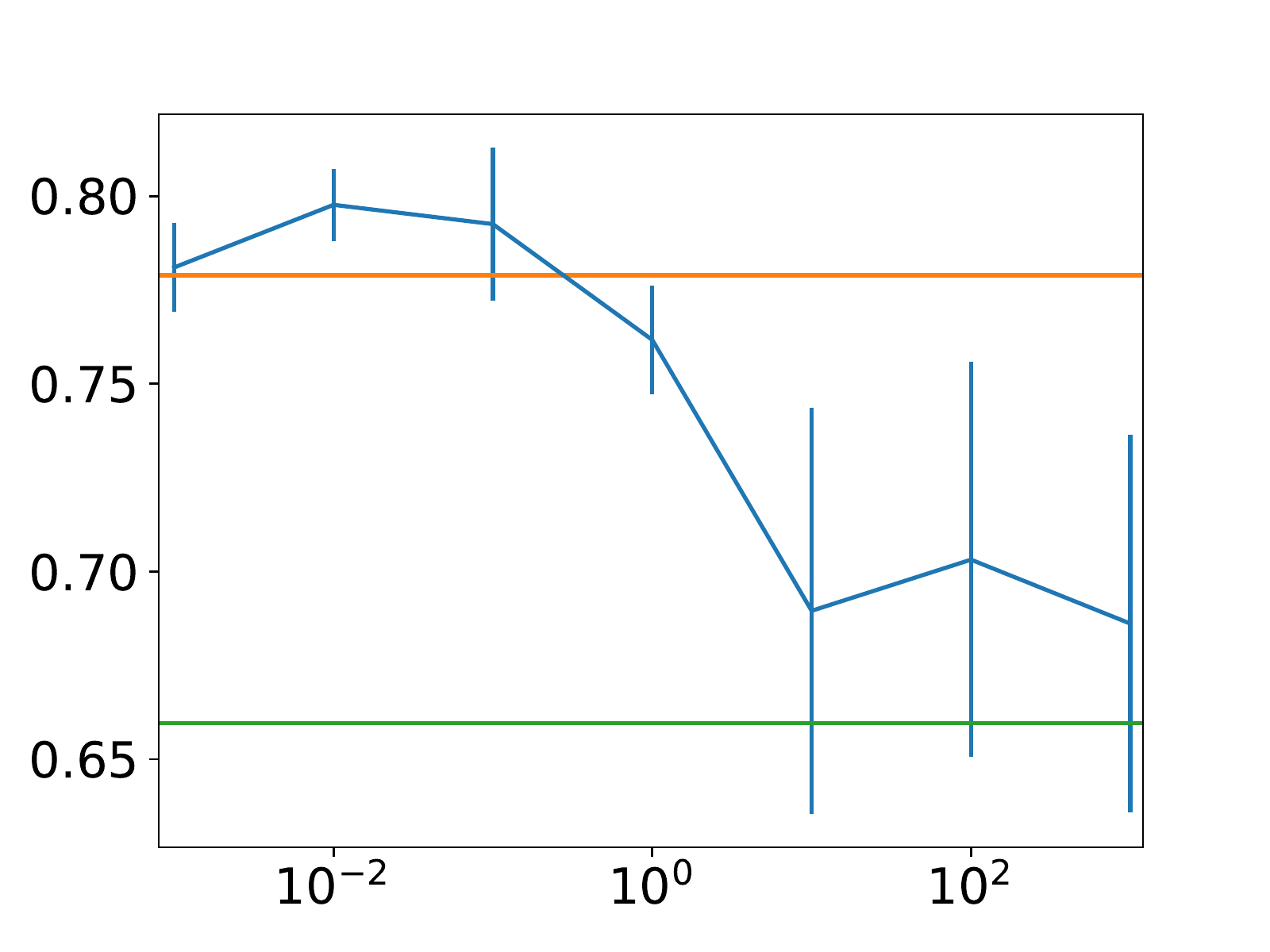} &
\includegraphics[width=11em]{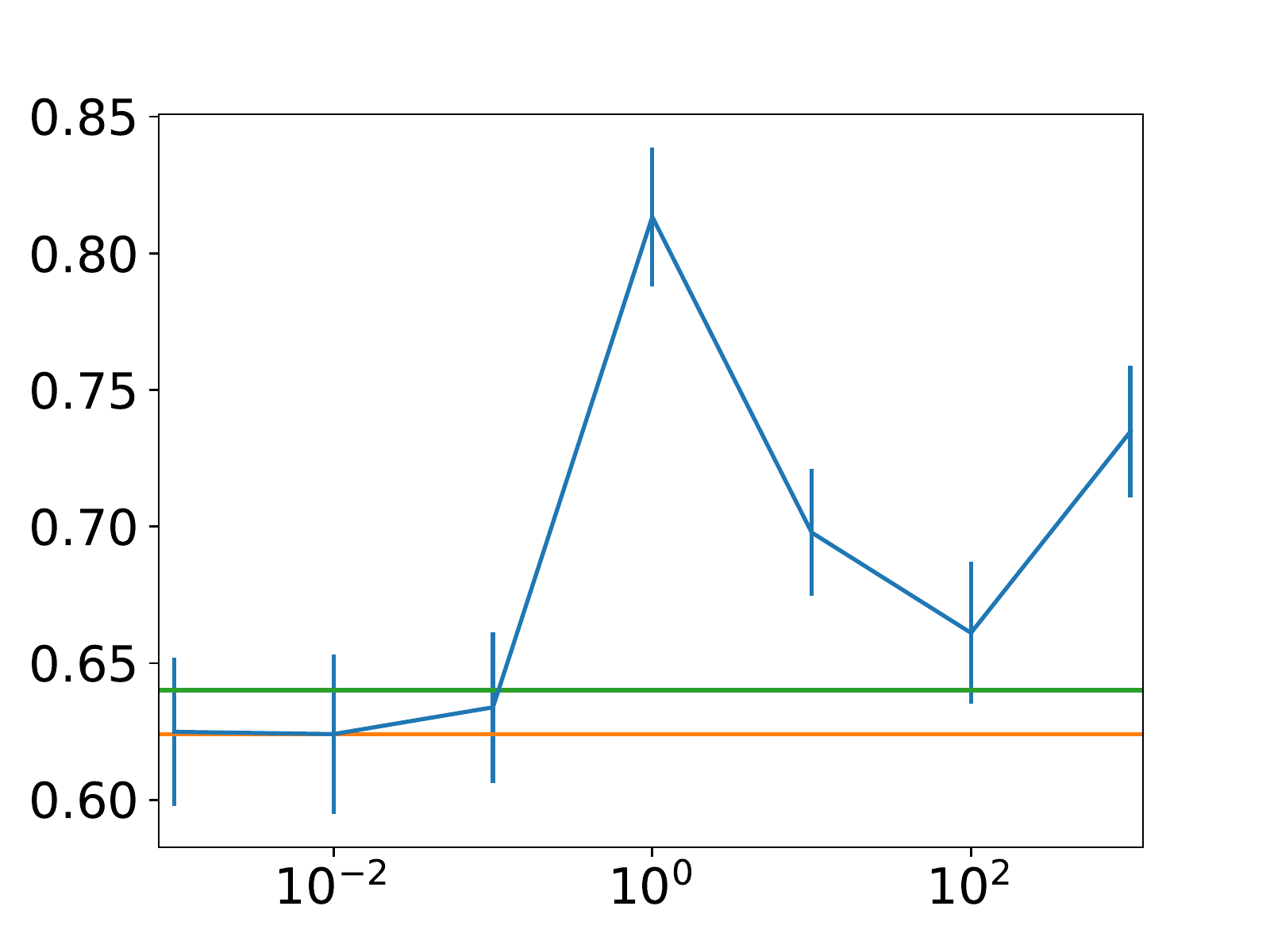} &
\includegraphics[width=11em]{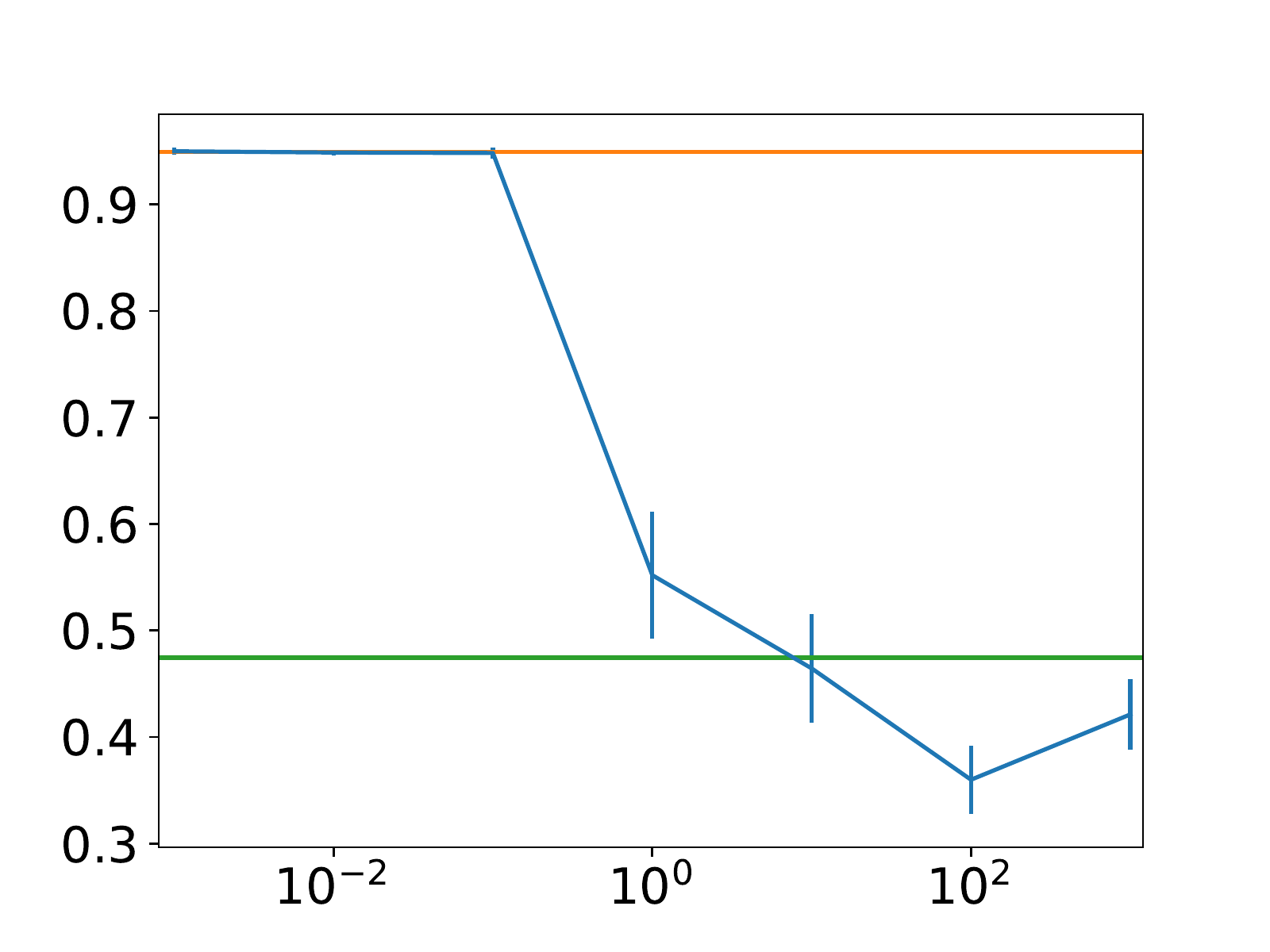} \\
SpamBase&
Waveform&
Wilt\\
   \end{tabular}
    \caption{AUC by the proposed method with different hyperparameters $\lambda$ trained on data with ten inexact anomaly sets with five instances per set. The x-axis is hyperparameter $\lambda$ and the y-axis is AUC. Errorbar shows standard error. The green and orange horizontal lines are AUC by the proposed method with $\lambda=\infty$ and $\lambda=0$, respectively.}
    \label{fig:lambda_auc}
   \end{figure*}

Figure~\ref{fig:lambda_auc} shows test AUC on the nine anomaly detection datasets
by the proposed method with different hyperparameters $\lambda$.
The best hyperparameters were different across datasets.
For example,
a high $\lambda$ was better with the Pima dataset,
a low $\lambda$ was better with the PageBlocks and Wilt datasets, and
an intermediate $\lambda$ was better with the Annthyroid and Waveform datasets.
The proposed method achieved high performance with various datasets
by automatically adapting $\lambda$ using the validation data
to control the balance of the anomaly score minimization for non-anomalous instances and inexact AUC maximization.

\section{Conclusion}
\label{sec:conclusion}

We proposed an extension of the AUC for inexact labels,
and developed a supervised anomaly detection method for data with inexact labels.
With our proposed method, we trained a neural network-based anomaly score function
by maximizing the inexact AUC while minimizing the anomaly scores for non-anomalous instances.
We experimentally confirmed its effectiveness using various datasets.
For future work, 
we would like to extend our framework for
semi-supervised settings~\citep{blanchard2010semi},
where unlabeled instances, labeled anomalous and labeled non-anomalous instances are given for training.

\bibliographystyle{abbrvnat}
\bibliography{ml2019}

\begin{thebibliography}{63}
\providecommand{\natexlab}[1]{#1}
\providecommand{\url}[1]{\texttt{#1}}
\expandafter\ifx\csname urlstyle\endcsname\relax
  \providecommand{\doi}[1]{doi: #1}\else
  \providecommand{\doi}{doi: \begingroup \urlstyle{rm}\Url}\fi

\bibitem[Akcay et~al.(2018)Akcay, Atapour-Abarghouei, and
  Breckon]{akcay2018ganomaly}
S.~Akcay, A.~Atapour-Abarghouei, and T.~P. Breckon.
\newblock Ganomaly: Semi-supervised anomaly detection via adversarial training.
\newblock In \emph{14th Asian Conference on Computer Vision}, 2018.

\bibitem[Aleskerov et~al.(1997)Aleskerov, Freisleben, and
  Rao]{aleskerov1997cardwatch}
E.~Aleskerov, B.~Freisleben, and B.~Rao.
\newblock Cardwatch: A neural network based database mining system for credit
  card fraud detection.
\newblock In \emph{IEEE/IAFE Computational Intelligence for Financial
  Engineering}, pages 220--226, 1997.

\bibitem[An and Cho(2015)]{an2015variational}
J.~An and S.~Cho.
\newblock Variational autoencoder based anomaly detection using reconstruction
  probability.
\newblock \emph{Special Lecture on IE}, 2:\penalty0 1--18, 2015.

\bibitem[Andrews et~al.(2003)Andrews, Tsochantaridis, and
  Hofmann]{andrews2003support}
S.~Andrews, I.~Tsochantaridis, and T.~Hofmann.
\newblock Support vector machines for multiple-instance learning.
\newblock In \emph{Advances in Neural Information Processing Systems}, pages
  577--584, 2003.

\bibitem[Babenko et~al.(2009)Babenko, Yang, and Belongie]{babenko2009visual}
B.~Babenko, M.-H. Yang, and S.~Belongie.
\newblock Visual tracking with online multiple instance learning.
\newblock In \emph{IEEE Conference on Computer Vision and Pattern Recognition},
  pages 983--990. IEEE, 2009.

\bibitem[Blanchard et~al.(2010)Blanchard, Lee, and Scott]{blanchard2010semi}
G.~Blanchard, G.~Lee, and C.~Scott.
\newblock Semi-supervised novelty detection.
\newblock \emph{Journal of Machine Learning Research}, 11\penalty0
  (Nov):\penalty0 2973--3009, 2010.

\bibitem[Brefeld and Scheffer(2005)]{brefeld2005auc}
U.~Brefeld and T.~Scheffer.
\newblock {AUC} maximizing support vector learning.
\newblock In \emph{Proceedings of the ICML Workshop on ROC Analysis in Machine
  Learning}, 2005.

\bibitem[Breiman(2001)]{breiman2001random}
L.~Breiman.
\newblock Random forests.
\newblock \emph{Machine learning}, 45\penalty0 (1):\penalty0 5--32, 2001.

\bibitem[Breunig et~al.(2000)Breunig, Kriegel, Ng, and Sander]{breunig2000lof}
M.~M. Breunig, H.-P. Kriegel, R.~T. Ng, and J.~Sander.
\newblock {LOF}: identifying density-based local outliers.
\newblock \emph{ACM SIGMOD Record}, 29\penalty0 (2):\penalty0 93--104, 2000.

\bibitem[Campos et~al.(2016)Campos, Zimek, Sander, Campello, Micenkov{\'a},
  Schubert, Assent, and Houle]{campos2016evaluation}
G.~O. Campos, A.~Zimek, J.~Sander, R.~J. Campello, B.~Micenkov{\'a},
  E.~Schubert, I.~Assent, and M.~E. Houle.
\newblock On the evaluation of unsupervised outlier detection: measures,
  datasets, and an empirical study.
\newblock \emph{Data Mining and Knowledge Discovery}, 30\penalty0 (4):\penalty0
  891--927, 2016.

\bibitem[Carbonneau et~al.(2018)Carbonneau, Cheplygina, Granger, and
  Gagnon]{carbonneau2018multiple}
M.-A. Carbonneau, V.~Cheplygina, E.~Granger, and G.~Gagnon.
\newblock Multiple instance learning: A survey of problem characteristics and
  applications.
\newblock \emph{Pattern Recognition}, 77:\penalty0 329--353, 2018.

\bibitem[Chandola et~al.(2009)Chandola, Banerjee, and
  Kumar]{chandola2009anomaly}
V.~Chandola, A.~Banerjee, and V.~Kumar.
\newblock Anomaly detection: A survey.
\newblock \emph{ACM Computing Surveys}, 41\penalty0 (3):\penalty0 15, 2009.

\bibitem[Chong and Tay(2017)]{chong2017abnormal}
Y.~S. Chong and Y.~H. Tay.
\newblock Abnormal event detection in videos using spatiotemporal autoencoder.
\newblock In \emph{International Symposium on Neural Networks}, pages 189--196.
  Springer, 2017.

\bibitem[Cinbis et~al.(2017)Cinbis, Verbeek, and Schmid]{cinbis2017weakly}
R.~G. Cinbis, J.~Verbeek, and C.~Schmid.
\newblock Weakly supervised object localization with multi-fold multiple
  instance learning.
\newblock \emph{IEEE Transactions on Pattern Analysis and Machine
  Intelligence}, 39\penalty0 (1):\penalty0 189--203, 2017.

\bibitem[Cortes and Mohri(2004)]{cortes2004auc}
C.~Cortes and M.~Mohri.
\newblock {AUC} optimization vs. error rate minimization.
\newblock In \emph{Advances in Neural Information Processing Systems}, pages
  313--320, 2004.

\bibitem[Das et~al.(2016)Das, Wong, Dietterich, Fern, and
  Emmott]{das2016incorporating}
S.~Das, W.-K. Wong, T.~Dietterich, A.~Fern, and A.~Emmott.
\newblock Incorporating expert feedback into active anomaly discovery.
\newblock In \emph{16th International Conference on Data Mining}, pages
  853--858. IEEE, 2016.

\bibitem[Das et~al.(2017)Das, Wong, Fern, Dietterich, and
  Siddiqui]{das2017incorporating}
S.~Das, W.-K. Wong, A.~Fern, T.~G. Dietterich, and M.~A. Siddiqui.
\newblock Incorporating feedback into tree-based anomaly detection.
\newblock In \emph{KDD Workshop on Interactive Data Exploration and Analytics},
  2017.

\bibitem[Dietterich et~al.(1997)Dietterich, Lathrop, and
  Lozano-P{\'e}rez]{dietterich1997solving}
T.~G. Dietterich, R.~H. Lathrop, and T.~Lozano-P{\'e}rez.
\newblock Solving the multiple instance problem with axis-parallel rectangles.
\newblock \emph{Artificial intelligence}, 89\penalty0 (1-2):\penalty0 31--71,
  1997.

\bibitem[Dodd and Pepe(2003)]{dodd2003partial}
L.~E. Dodd and M.~S. Pepe.
\newblock Partial {AUC} estimation and regression.
\newblock \emph{Biometrics}, 59\penalty0 (3):\penalty0 614--623, 2003.

\bibitem[Dokas et~al.(2002)Dokas, Ertoz, Kumar, Lazarevic, Srivastava, and
  Tan]{dokas2002data}
P.~Dokas, L.~Ertoz, V.~Kumar, A.~Lazarevic, J.~Srivastava, and P.-N. Tan.
\newblock Data mining for network intrusion detection.
\newblock In \emph{NSF Workshop on Next Generation Data Mining}, pages 21--30,
  2002.

\bibitem[Eskin(2000)]{eskin2000anomaly}
E.~Eskin.
\newblock Anomaly detection over noisy data using learned probability
  distributions.
\newblock In \emph{International Conference on Machine Learning}, 2000.

\bibitem[Feng and Zhou(2017)]{feng2017deep}
J.~Feng and Z.-H. Zhou.
\newblock Deep miml network.
\newblock In \emph{Thirty-First AAAI Conference on Artificial Intelligence},
  2017.

\bibitem[Fujimaki et~al.(2005)Fujimaki, Yairi, and
  Machida]{fujimaki2005approach}
R.~Fujimaki, T.~Yairi, and K.~Machida.
\newblock An approach to spacecraft anomaly detection problem using kernel
  feature space.
\newblock In \emph{International Conference on Knowledge Discovery in Data
  Mining}, pages 401--410, 2005.

\bibitem[Fujino and Ueda(2016)]{fujino2016semi}
A.~Fujino and N.~Ueda.
\newblock A semi-supervised {AUC} optimization method with generative models.
\newblock In \emph{16th International Conference on Data Mining}, pages
  883--888. IEEE, 2016.

\bibitem[Gao et~al.(2006)Gao, Cheng, and Tan]{gao2006novel}
J.~Gao, H.~Cheng, and P.-N. Tan.
\newblock A novel framework for incorporating labeled examples into anomaly
  detection.
\newblock In \emph{Proceedings of the 2006 SIAM International Conference on
  Data Mining}, pages 594--598. SIAM, 2006.

\bibitem[Hanley and McNeil(1982)]{hanley1982meaning}
J.~A. Hanley and B.~J. McNeil.
\newblock The meaning and use of the area under a receiver operating
  characteristic ({ROC}) curve.
\newblock \emph{Radiology}, 143\penalty0 (1):\penalty0 29--36, 1982.

\bibitem[Herrera et~al.(2016)Herrera, Ventura, Bello, Cornelis, Zafra,
  S{\'a}nchez-Tarrag{\'o}, and Vluymans]{herrera2016multiple}
F.~Herrera, S.~Ventura, R.~Bello, C.~Cornelis, A.~Zafra,
  D.~S{\'a}nchez-Tarrag{\'o}, and S.~Vluymans.
\newblock \emph{Multiple instance learning: foundations and algorithms}.
\newblock Springer, 2016.

\bibitem[Hodge and Austin(2004)]{hodge2004survey}
V.~Hodge and J.~Austin.
\newblock A survey of outlier detection methodologies.
\newblock \emph{Artificial Ntelligence Review}, 22\penalty0 (2):\penalty0
  85--126, 2004.

\bibitem[Id{\'e} and Kashima(2004)]{ide2004eigenspace}
T.~Id{\'e} and H.~Kashima.
\newblock Eigenspace-based anomaly detection in computer systems.
\newblock In \emph{International Conference on Knowledge Discovery and Data
  Mining}, pages 440--449, 2004.

\bibitem[Ilse et~al.(2018)Ilse, Tomczak, and Welling]{ilse2018attention}
M.~Ilse, J.~Tomczak, and M.~Welling.
\newblock Attention-based deep multiple instance learning.
\newblock In \emph{International Conference on Machine Learning}, pages
  2132--2141, 2018.

\bibitem[Iwata and Yamanaka(2019)]{iwata2019supervised}
T.~Iwata and Y.~Yamanaka.
\newblock Supervised anomaly detection based on deep autoregressive density
  estimators.
\newblock \emph{arXiv preprint arXiv:1904.06034}, 2019.

\bibitem[Kingma and Ba(2015)]{kingma2014adam}
D.~P. Kingma and J.~Ba.
\newblock {ADAM}: {A} method for stochastic optimization.
\newblock In \emph{International Conference on Learning Representations}, 2015.

\bibitem[Kingma and Wellniga(2014)]{kingma2013auto}
D.~P. Kingma and M.~Wellniga.
\newblock Auto-encoding variational {Bayes}.
\newblock In \emph{2nd International Conference on Learning Representations},
  2014.

\bibitem[Laxhammar et~al.(2009)Laxhammar, Falkman, and
  Sviestins]{laxhammar2009anomaly}
R.~Laxhammar, G.~Falkman, and E.~Sviestins.
\newblock Anomaly detection in sea traffic - a comparison of the {G}aussian
  mixture model and the kernel density estimator.
\newblock In \emph{International Conference on Information Fusion}, pages
  756--763, 2009.

\bibitem[Liu et~al.(2008)Liu, Ting, and Zhou]{liu2008isolation}
F.~T. Liu, K.~M. Ting, and Z.-H. Zhou.
\newblock Isolation forest.
\newblock In \emph{Proceeding of the 8th IEEE International Conference on Data
  Mining}, pages 413--422. IEEE, 2008.

\bibitem[Markou and Singh(2003)]{markou2003novelty}
M.~Markou and S.~Singh.
\newblock Novelty detection: a review.
\newblock \emph{Signal processing}, 83\penalty0 (12):\penalty0 2481--2497,
  2003.

\bibitem[Maron and Lozano-P{\'e}rez(1998)]{maron1998framework}
O.~Maron and T.~Lozano-P{\'e}rez.
\newblock A framework for multiple-instance learning.
\newblock In \emph{Advances in Neural Information Processing Systems}, pages
  570--576, 1998.

\bibitem[Mukkamala et~al.(2005)Mukkamala, Sung, and
  Ribeiro]{mukkamala2005model}
S.~Mukkamala, A.~Sung, and B.~Ribeiro.
\newblock Model selection for kernel based intrusion detection systems.
\newblock In \emph{Adaptive and Natural Computing Algorithms}, pages 458--461.
  Springer, 2005.

\bibitem[Munawar et~al.(2017)Munawar, Vinayavekhin, and
  De~Magistris]{Munawar_2017}
A.~Munawar, P.~Vinayavekhin, and G.~De~Magistris.
\newblock Limiting the reconstruction capability of generative neural network
  using negative learning.
\newblock In \emph{27th International Workshop on Machine Learning for Signal
  Processing}. IEEE, 2017.

\bibitem[Nadeem et~al.(2016)Nadeem, Marshall, Singh, Fang, and
  Yuan]{nadeem2016semi}
M.~Nadeem, O.~Marshall, S.~Singh, X.~Fang, and X.~Yuan.
\newblock Semi-supervised deep neural network for network intrusion detection.
\newblock In \emph{KSU Conference on Cybersecurity Education, Research and
  Practice}, 2016.

\bibitem[Narasimhan and Agarwal(2017)]{narasimhan2017support}
H.~Narasimhan and S.~Agarwal.
\newblock Support vector algorithms for optimizing the partial area under the
  {ROC} curve.
\newblock \emph{Neural Computation}, 29\penalty0 (7):\penalty0 1919--1963,
  2017.

\bibitem[Paszke et~al.(2017)Paszke, Gross, Chintala, Chanan, Yang, DeVito, Lin,
  Desmaison, Antiga, and Lerer]{paszke2017automatic}
A.~Paszke, S.~Gross, S.~Chintala, G.~Chanan, E.~Yang, Z.~DeVito, Z.~Lin,
  A.~Desmaison, L.~Antiga, and A.~Lerer.
\newblock Automatic differentiation in {PyTorch}.
\newblock In \emph{NIPS Autodiff Workshop}, 2017.

\bibitem[Patcha and Park(2007)]{patcha2007overview}
A.~Patcha and J.-M. Park.
\newblock An overview of anomaly detection techniques: Existing solutions and
  latest technological trends.
\newblock \emph{Computer Networks}, 51\penalty0 (12):\penalty0 3448--3470,
  2007.

\bibitem[Pedregosa et~al.(2011)Pedregosa, Varoquaux, Gramfort, Michel, Thirion,
  Grisel, Blondel, Prettenhofer, Weiss, Dubourg, et~al.]{pedregosa2011scikit}
F.~Pedregosa, G.~Varoquaux, A.~Gramfort, V.~Michel, B.~Thirion, O.~Grisel,
  M.~Blondel, P.~Prettenhofer, R.~Weiss, V.~Dubourg, et~al.
\newblock Scikit-learn: Machine learning in python.
\newblock \emph{Journal of Machine Learning Research}, 12:\penalty0 2825--2830,
  2011.

\bibitem[Pimentel et~al.(2018)Pimentel, Monteiro, Viana, Veloso, and
  Ziviani]{pimentel2018generalized}
T.~Pimentel, M.~Monteiro, J.~Viana, A.~Veloso, and N.~Ziviani.
\newblock A generalized active learning approach for unsupervised anomaly
  detection.
\newblock \emph{arXiv preprint arXiv:1805.09411}, 2018.

\bibitem[Pinheiro and Collobert(2015)]{pinheiro2015image}
P.~O. Pinheiro and R.~Collobert.
\newblock From image-level to pixel-level labeling with convolutional networks.
\newblock In \emph{Proceedings of the IEEE Conference on Computer Vision and
  Pattern Recognition}, pages 1713--1721, 2015.

\bibitem[Rapaka et~al.(2003)Rapaka, Novokhodko, and
  Wunsch]{rapaka2003intrusion}
A.~Rapaka, A.~Novokhodko, and D.~Wunsch.
\newblock Intrusion detection using radial basis function network on sequences
  of system calls.
\newblock In \emph{International Joint Conference on Neural Networks},
  volume~3, pages 1820--1825, 2003.

\bibitem[Sabokrou et~al.(2016)Sabokrou, Fathy, and Hoseini]{sabokrou2016video}
M.~Sabokrou, M.~Fathy, and M.~Hoseini.
\newblock Video anomaly detection and localisation based on the sparsity and
  reconstruction error of auto-encoder.
\newblock \emph{Electronics Letters}, 52\penalty0 (13):\penalty0 1122--1124,
  2016.

\bibitem[Sakai et~al.(2018)Sakai, Niu, and Sugiyama]{sakai2018semi}
T.~Sakai, G.~Niu, and M.~Sugiyama.
\newblock Semi-supervised {AUC} optimization based on positive-unlabeled
  learning.
\newblock \emph{Machine Learning}, 107\penalty0 (4):\penalty0 767--794, 2018.

\bibitem[Sakurada and Yairi(2014)]{sakurada2014anomaly}
M.~Sakurada and T.~Yairi.
\newblock Anomaly detection using autoencoders with nonlinear dimensionality
  reduction.
\newblock In \emph{Proceedings of the MLSDA 2nd Workshop on Machine Learning
  for Sensory Data Analysis}. ACM, 2014.

\bibitem[Sch{\"o}lkopf et~al.(2001)Sch{\"o}lkopf, Platt, Shawe-Taylor, Smola,
  and Williamson]{scholkopf2001estimating}
B.~Sch{\"o}lkopf, J.~C. Platt, J.~Shawe-Taylor, A.~J. Smola, and R.~C.
  Williamson.
\newblock Estimating the support of a high-dimensional distribution.
\newblock \emph{Neural Computation}, 13\penalty0 (7):\penalty0 1443--1471,
  2001.

\bibitem[Sch{\"o}lkopf et~al.(2002)Sch{\"o}lkopf, Smola,
  et~al.]{scholkopf2002learning}
B.~Sch{\"o}lkopf, A.~J. Smola, et~al.
\newblock \emph{Learning with kernels: support vector machines, regularization,
  optimization, and beyond}.
\newblock MIT press, 2002.

\bibitem[Shewhart(1931)]{shewhart1931economic}
W.~A. Shewhart.
\newblock \emph{Economic control of quality of manufactured product}.
\newblock ASQ Quality Press, 1931.

\bibitem[Singh and Silakari(2009)]{singh2009ensemble}
S.~Singh and S.~Silakari.
\newblock An ensemble approach for feature selection of cyber attack dataset.
\newblock \emph{arXiv preprint arXiv:0912.1014}, 2009.

\bibitem[Suh et~al.(2016)Suh, Chae, Kang, and Choi]{suh2016echo}
S.~Suh, D.~H. Chae, H.-G. Kang, and S.~Choi.
\newblock Echo-state conditional variational autoencoder for anomaly detection.
\newblock In \emph{International Joint Conference on Neural Networks}, pages
  1015--1022, 2016.

\bibitem[Wong et~al.(2003)Wong, Moore, Cooper, and Wagner]{wong2003bayesian}
W.-K. Wong, A.~W. Moore, G.~F. Cooper, and M.~M. Wagner.
\newblock Bayesian network anomaly pattern detection for disease outbreaks.
\newblock In \emph{International Conference on Machine Learning}, pages
  808--815, 2003.

\bibitem[Wu et~al.(2015)Wu, Yu, Huang, and Yu]{wu2015deep}
J.~Wu, Y.~Yu, C.~Huang, and K.~Yu.
\newblock Deep multiple instance learning for image classification and
  auto-annotation.
\newblock In \emph{Proceedings of the IEEE Conference on Computer Vision and
  Pattern Recognition}, pages 3460--3469, 2015.

\bibitem[Xu et~al.(2018)Xu, Chen, Zhao, Li, Bu, Li, Liu, Zhao, Pei, Feng,
  et~al.]{xu2018unsupervised}
H.~Xu, W.~Chen, N.~Zhao, Z.~Li, J.~Bu, Z.~Li, Y.~Liu, Y.~Zhao, D.~Pei, Y.~Feng,
  et~al.
\newblock Unsupervised anomaly detection via variational auto-encoder for
  seasonal kpis in web applications.
\newblock In \emph{World Wide Web Conference}, pages 187--196, 2018.

\bibitem[Yamanishi et~al.(2004)Yamanishi, Takeuchi, Williams, and
  Milne]{yamanishi2004line}
K.~Yamanishi, J.-I. Takeuchi, G.~Williams, and P.~Milne.
\newblock On-line unsupervised outlier detection using finite mixtures with
  discounting learning algorithms.
\newblock \emph{Data Mining and Knowledge Discovery}, 8\penalty0 (3):\penalty0
  275--300, 2004.

\bibitem[Ying et~al.(2016)Ying, Wen, and Lyu]{ying2016stochastic}
Y.~Ying, L.~Wen, and S.~Lyu.
\newblock Stochastic online {AUC} maximization.
\newblock In \emph{Advances in Neural Information Processing Systems}, pages
  451--459, 2016.

\bibitem[Zhai et~al.(2016)Zhai, Cheng, Lu, and Zhang]{zhai2016deep}
S.~Zhai, Y.~Cheng, W.~Lu, and Z.~Zhang.
\newblock Deep structured energy based models for anomaly detection.
\newblock In \emph{International Conference on Machine Learning}, pages
  1100--1109, 2016.

\bibitem[Zhou and Paffenroth(2017)]{zhou2017anomaly}
C.~Zhou and R.~C. Paffenroth.
\newblock Anomaly detection with robust deep autoencoders.
\newblock In \emph{Proceedings of the 23rd ACM SIGKDD International Conference
  on Knowledge Discovery and Data Mining}, pages 665--674. ACM, 2017.

\bibitem[Zhu et~al.(2017)Zhu, Lou, Vang, and Xie]{zhu2017deep}
W.~Zhu, Q.~Lou, Y.~S. Vang, and X.~Xie.
\newblock Deep multi-instance networks with sparse label assignment for whole
  mammogram classification.
\newblock In \emph{International Conference on Medical Image Computing and
  Computer-Assisted Intervention}, pages 603--611, 2017.

\end{thebibliography}

\end{document}